%% file: main_arxiv.tex
\documentclass[runningheads]{llncs}

\usepackage{eccv}

\usepackage{eccvabbrv}

\usepackage{graphicx}
\usepackage{booktabs}
\usepackage{placeins}
\usepackage{listings}

\newcommand{\imgcopyright}{\par\vspace{2pt}
  {\scriptsize Images \copyright{} Inter IKEA Systems~B.V.~2026, used with permission\par}}
\usepackage[accsupp]{axessibility}  

\usepackage{hyperref}

\usepackage{orcidlink}

\begin{document}

\title{PolyLayout: Hierarchical VLM-Guided Layout Generation Beyond Rectangular Rooms} 

\titlerunning{PolyLayout}

\author{Yutong Jiang \and
Zahra Atashgahi \and
Carlos Soto Garcia Delgado \and
Ruben Brokkelkamp \and
Davide Zanutto \and
Ef\c{s}an S\"okmen \and
Shahin Shahkarami}

\authorrunning{Y.~Jiang et al.}

\institute{IKEA Retail (Ingka Group), Amsterdam, The Netherlands\\
\email{\{yutong.jiang2,efsan.sokmen\}@ingka.ikea.com}}

\maketitle

\input{sections/abstract}
\input{sections/introduction}

\input{sections/related_work}
\input{sections/methods}

\input{sections/fine_placement}
\input{sections/experiments}

\input{sections/conclusion}

\section*{Acknowledgements}
We thank our colleagues in the Home Imagination Platform team at IKEA Retail (Ingka Group) for their
support, especially Thilek Silvadorai, Chris Root, as well as Bassem Samir, and Can G\"o\c{c}meno\u{g}lu
from Geomagical Labs for their help with room builder and rendering
pipeline.
All furniture assets shown in figures appear courtesy of Inter IKEA Systems~B.V. The IKEA logo and the IKEA
wordmark are registered trademarks of Inter IKEA Systems~B.V.

%
%
\bibliographystyle{splncs04}
\bibliography{main}

\clearpage
\appendix
\section*{Supplementary Material}

This supplementary material provides the full prompt templates of the
\textsc{PolyLayout} pipeline (Appendix~\ref{supp:stage2}) and of the VLM
plausibility judge used in our evaluation (Appendix~\ref{supp:judge}). Curly-braced
fields (\eg, \texttt{\{room\_type\}}) are template variables filled at
runtime with the room context described in the main paper.

\section{Stage-2 Prompt Templates}
\label{supp:stage2}

Stage~2 of the pipeline (Sec.~3.1 of the main paper) issues four VLM calls
in sequence: placement ordering, RAG-based context planning, semantic
suggestion, and structured extraction. A preprocessing call classifies
wall-mounted items. The exact templates follow.

\subsection{Placement Ordering}
\label{supp:ordering}
Determines the global placement order over functional bundles and
standalone items (``Placement Ordering'' in the main paper). Bundles are
referred to as \emph{bubbles} in the implementation.
\begin{lstlisting}
<instruction>
You are an expert interior designer. Your task is to determine the placement priority for a list of furniture items and functional bubbles to be placed in a {room_type}.

The room may already contain placed items (bubbles and single furniture). These placed elements are provided only for context. You must prioritize ONLY the elements that still need to be placed.

The placement system works iteratively: it places the most important items first. If an item cannot be placed, it is skipped. Therefore, the order you define is critical to ensure the most essential elements are included in the final layout.

Your goal is to order all to-place elements from most important to least important, ensuring functional balance for the room.
</instruction>

<context>
----------------
ROOM INFORMATION
----------------
- Room Type: {room_type}
- Walls: {walls_dimensions_query}
- Window(s): {window_information}
- Door(s): {door_information}

----------------
PLACED ELEMENTS (CONTEXT ONLY)
----------------
- Placed Functional Bubbles: {placed_formatted_bubble_list}
- Placed Single Furniture Items: {placed_formatted_single_furniture_list}

----------------
ELEMENTS TO ORDER (TO PLACE)
----------------
- Functional Bubbles: {to_place_formatted_bubble_list}
- Single Furniture Items: {to_place_single_furniture_list}
</context>

<output_format>
Return a single, flat Python list containing ONLY the names of all to-place bubbles and to-place single furniture items in their prioritized order. The list should contain nothing but the string names of the to-place elements.
</output_format>

<example>
# Input (Empty Room Case):
# placed_bubble_list: []
# placed_furniture_list: []
# to_place_bubble_list: ["sleeping area_1"]
# to_place_furniture_list: ["rug_1", "nightstand_1", "wardrobe_1", "bed_2"]
# room_type: "Bedroom"

# Output:
["sleeping area_1", "wardrobe_1", "nightstand_1", "rug_1", "bed_2"]
</example>

<example>
# Input (Partially Furnished Case):
# placed_bubble_list: ["resting_area_1"]
# placed_furniture_list: ["tv_unit_1"]
# to_place_bubble_list: ["eating_area_1"]
# to_place_furniture_list: ["table_with_four_chairs_1", "three_place_sofa_1", "coffee_table_1", "rug_1", "floor_lamp_1", "drawer_1", "beige_sofa_1", "beige_sofa_2", "frame_1", "frame_2"]
# room_type: "Small Living-Dining Room"

# Output:
["eating_area_1",
 "table_with_four_chairs_1",
 "three_place_sofa_1",
 "rug_1",
 "coffee_table_1",
 "floor_lamp_1",
 "drawer_1",
 "beige_sofa_1",
 "beige_sofa_2",
 "frame_1",
 "frame_2"]
</example>

<instruction>
Now, generate the prioritized Python list for ONLY the to-place elements provided.
Use the placed elements as context to maintain functional balance but do not include them in the output list.
</instruction>
\end{lstlisting}

\subsection{Context Plan via Retrieval-Augmented Generation}
\label{supp:plan}
Synthesizes the retrieved interior-design guidelines
(\texttt{\{context\_guidelines\}}) into a high-level spatial plan
(``Context Plan via Retrieval-Augmented Generation'' in the main paper).
\begin{lstlisting}
<instruction>
You are a module in charge of generating a high level plan about how to place furniture in a room.

REFERENCE OBJECT DEFINITIONS
Walls: [{wall_list}]
Windows: [{window_list}]
Doors: [{door_list}]
Room type: {room_type}

The room may already contain placed elements (bubbles and single furniture). These are context and should influence how you plan the placement of the remaining to-place elements.

Elements to consider:
- Placed (Context Only)
  - Bubbles: {placed_formatted_bubble_list}
  - Single Furniture: {placed_formatted_single_furniture_list}
- To Place (Plan for these)
  - Bubbles: {to_place_formatted_bubble_list}
  - Single Furniture: {to_place_single_furniture_list}

Elements have an order of importance defined in the following list: {ordering_list}

Wall information: {walls_dimensions_query}
Window information: {window_information}
Door information: {door_information}

Your goal is to create a high level plan that outlines how to place the TO-PLACE furniture into the room given the context of already placed elements.
Do not provide specific placements, numeric values or function calls. Focus on the overall strategy and considerations for arranging the to-place furniture effectively into the provided room.
You need to provide a plan for all TO-PLACE furniture objects (bubbles and single furniture). Refer to each bubble as a whole, not individual furniture within the bubble.

CRITICAL — DOOR & WINDOW CLEARANCE:
- The image shows doors (red) and windows (blue) with their positions clearly marked. The room dimensions (wall lengths and distances) are also annotated.
- Use spatial awareness: consider the size of furniture relative to the available wall space between openings.
- NEVER place furniture that blocks or overlaps with a door's swing area or passage zone.
- NEVER place furniture directly in front of a window in a way that blocks access or overlaps with the window opening area.
- Treat the full width of each door and window as a no-furniture zone.
- When placing furniture "next to" a window or door, ensure it is beside the opening, not covering it.
- Be mindful of how much free wall space remains after accounting for doors and windows — do not overcrowd narrow wall segments.

For context, consider the following guidelines for placing furniture:
{context_guidelines}
{layout_style_directive}

IMPORTANT OUTPUT FORMAT:
- Keep your response CONCISE: maximum 300 words.
- Use brief bullet points, not lengthy paragraphs.
- State WHICH bubble goes on WHICH wall/corner, and WHY in one sentence.
- Do NOT repeat the room geometry or furniture dimensions back.
- Do NOT write section headers, markdown formatting, or numbered sub-sections.
\end{lstlisting}

\subsection{Semantic Suggestion}
\label{supp:suggestion}
The multi-modal placement prompt (``Semantic Suggestion'' in the main
paper). It accompanies the annotated top-down rendering of the room and
produces one free-text placement rationale per element.
\begin{lstlisting}
<instruction>

You are an interior designer tasked with placing furniture items in a room.

----------------
ROOM CONTEXT
----------------
1. Room Context
Room Type: {room_type}
Nr. of Walls: {nr_walls}
Walls: {walls_dimensions_query}
Window(s): {window_information}
Door(s): {door_information}

----------------
CURRENT LAYOUT
----------------
2. Current Layout
Top-Down View:
You are provided with a top-down image of the room showing its current state.

Furniture can come in two different forms:
-Single furniture item: A single piece of furniture
-Bubble: A group of single furniture items with a fixed position relative positions

Furniture items to place: {furniture_list}
Functional Bubbles to Place: {bubble_list}

- Relative positions within each bubble: {relative_pos_store}

Elements have an order of importance defined in the following list: {ordering_list}

Your task is to suggest placements for **each bubble and each single furniture item in the single furniture items list** based on the room layout and existing furniture.
<context>
Another expert has provided placement suggestions that you can use as context: {context_plan}
</context>

</instruction>

<constraint>
Follow these design and placement rules:

1. Base all placements on room dimensions, existing furniture, and good design principles.
2. If multiple elements are provided, place them in a logical sequence. Place large elements first then smaller elements. If you want to position one object relative to another, make sure the reference object is placed earlier in the order.
3. For each element, provide a clear placement explanation wrapped in `<reas>...</reas>` tags **only**.
4. You can reference:
   - Existing or previously placed elements
   -- Room itself: "room"
   - Wall: [{wall_list}]
   - Window: [{window_list}]
   - Door: [{door_list}]
5. Valid positions relative to the **reference object**:
   - If reference object is existing or previously placed furniture: the reference position should be chosen from ["in front", "left side", "right side", "on", "under"].
   - If reference object is wall: the reference position should be chosen from ["center", "hang"]. NOTE: "hang" is ONLY for wall-mounted items. Floor furniture MUST use "center".
   - If reference object is room: the reference position should be chosen from [{corner_list}]
   - If reference object is window: reference position should be chosen from ["next to"]
   - If reference object is door: reference position should be chosen from ["next to"]
5a. Disambiguation of "left side" and "right side":
    - Imagine traversing the room perimeter in a clockwise direction.
    - When you reach (conceptually align with) a referenced furniture item, the side of that item you would encounter FIRST in that clockwise traversal is its "right side".
    - The opposite lateral side (encountered second) is its "left side".
    - Always apply this clockwise rule; do not use a viewer-centric or camera perspective.

Wall center vs corner leverage:
- Choosing a wall "center" fixes the element at the midpoint, clarifying which wall segment is occupied.
- Choosing a room corner places the element squarely in that corner, reserving adjacent wall centers for other functions.
Use this deliberately to optimize spatial distribution and avoid congestion.

6. If furniture to be placed is a large piece, it should be placed with respect to the wall or room, not with respect to other furniture.
7. Be precise and avoid vague spatial terms. Only one reference object is allowed per placement.
8. Remember 'next to' could only be used when the reference object is a door or window!
9. Wall-mounted items MUST use a wall as their reference object with "hang" as the position. They CANNOT use other furniture as reference or use "on" relationship.

OVERLAP PREVENTION: Assign a unique placement to every furniture item. Identical placements cause items to occupy the same space and overlap. You must ensure all item placements are distinct.

DOOR & WINDOW CLEARANCE: Do NOT place any furniture that overlaps with or blocks a door or window. Doors need their full swing/passage area kept clear. Windows should not be obstructed by furniture placed directly in front of them. The image shows doors (red) and windows (blue) — use the room dimensions to judge available space and ensure furniture does not encroach on these openings.

BUBBLE INTEGRITY: A bubble is an atomic unit. Do NOT generate separate `<reas>` blocks for individual furniture items contained within a bubble. Only generate one `<reas>` per bubble as a whole. The individual items inside a bubble (listed in its description) are placed automatically relative to the bubble — you must not place them independently.
</constraint>

----------------
WALL-MOUNTED ITEMS
----------------
The following items MUST be hung on a wall. For each, select the most appropriate wall based on the room layout, existing furniture, and design principles. These items cannot be placed on the floor or on other furniture.
Wall-mounted items: {wall_furniture_list}

For each wall-mounted item, provide a `<reas>` block explaining your wall choice, then in the extraction step use relationship "hang" with the chosen wall.

<example>
<reas>
Hanging the frame on the pink wall above the sofa creates a focal point and balances the room visually.
</reas>
</example>

<instruction>
For each element in {furniture_list} and in {bubble_list}, do the following:
1. Choose an appropriate placement relative to one valid reference.
2. Justify the placement in a single `<reas>...</reas>` block.

For each wall-mounted item in the wall-mounted items list, do the following:
1. Choose the most appropriate wall to hang it on.
2. Justify the wall choice in a single `<reas>...</reas>` block.

IMPORTANT: Generate exactly one `<reas>` per bubble, one `<reas>` per single furniture item, and one `<reas>` per wall-mounted item. Do NOT generate `<reas>` blocks for individual furniture items inside a bubble.

Repeat for all elements.

OUTPUT FORMAT REQUIREMENT:
- Your output MUST consist ONLY of `<reas>...</reas>` blocks, one per element.
- Do NOT output JSON, numbered lists, markdown, or any other format.
- Do NOT wrap your response in ```json or ``` code fences.
- Each `<reas>` block should contain a brief placement explanation referencing the chosen wall/corner/furniture.
</instruction>
\end{lstlisting}

\subsection{Structured Extraction}
\label{supp:extraction}
Parses the semantic suggestions into executable
\texttt{place\_object} calls, enforcing the admissibility rules described
in the main paper (``Structured Extraction'').
\begin{lstlisting}
<instruction>
You are generating structured placement functions for both individual furniture items and functional bubbles based on provided suggestions.

REFERENCE OBJECT DEFINITIONS
Walls: [{wall_list}]
Windows: [{window_list}]
Doors: [{door_list}]
Room: "room"

The room contains the following elements to be placed:
- Bubbles: {bubble_list}
- Single Furniture: {furniture_list}
- Wall-Mounted Items (must be hung on a wall): {wall_furniture_list}

Wall information: {walls_dimensions_query}
Window information: {window_information}
Door information: {door_information}

For each placement suggestion, extract the necessary information and format it using the function call shown below.
Do **not** include any reasoning, explanation, or extra text in your output.
</instruction>

<constraint>
Follow these formatting rules and placement constraints:

1. Use the exact format below for each element placement:
   <func> place_object("element name", [length, height, width], ["reference object"], ["relative position"]) </func>
2. Valid reference objects:
   - Previously placed furniture or bubbles (names must exactly match those from the input lists or suggestions)
   - Walls: [{wall_list}]
   - Room: "room"
   - Windows: [{window_list}]
   - Doors: [{door_list}]
3. Valid relative positions based on reference type:
   - If the reference object is furniture or a bubble: ["in front", "left side", "right side", "on", "under"] (Note: This is typically for single furniture, not bubbles).
   - If the reference object is room: [{corner_list}]
   - If the reference object is wall: ["center", "hang"]
   - If the reference object is window: ["next to"]
   - If the reference object is door: ["next to"]
   NOTE: "hang" is ONLY for wall-mounted items. Floor furniture uses "center" for walls.
3a. Disambiguation of "left side" and "right side":
    - Assume a clockwise traversal of the room perimeter.
    - When conceptually reaching a referenced furniture item during this clockwise traversal, the side encountered first is its "right side"; the opposite lateral side is its "left side".
    - Always apply this rule; do not infer viewer-facing orientation.
4. If the element size is not specified, use: [0, 0, 0]
5. Use only one reference object per function call.
6. Remember 'next to' could only be used when the reference object is a door or window!
7. If the suggestion contains invalid phrasing, correct it:
   - Example: "center of red wall" → "center", "red wall"
8. Do not include any extra commentary or formatting outside the <func>...</func> blocks.
9. Corner positions in {corner_list} are only allowed when the reference object is "room".
# (Removed: The value "center" is also only allowed when the reference object is "room".)
10. Door Placement rules:
   - ANY wall that contains a door CANNOT have an element placed at the "center" of that wall.
   - If a suggestion places an element at the "center" of a wall with a door, you MUST replace the placement to use "next to" the door as the position, and the door as the reference object.
11. Bubble Placement Rules:
    - Bubbles can only be placed relative to the room, a wall, a window, or a door. They CANNOT be placed relative to other furniture or bubbles.
    - If a suggestion places a bubble relative to another furniture item, re-evaluate and place it relative to the nearest valid wall, corner, window, or door.
12. Uniqueness for Bubbles:
   - Only one bubble may occupy a specific wall center.
   - Only one bubble may occupy a specific room corner.
   - If a duplicate occurs, reassign the later bubble to the first available unused valid location (prefer unused wall centers on walls without doors, then unused room corners).
13. Bubble Integrity:
   - A bubble is an atomic unit. Do NOT generate separate `<func>` calls for individual furniture items contained within a bubble.
   - Only generate one `<func>` per bubble as a whole. The individual items inside a bubble are positioned automatically.
14. Wall-Mounted Items:
   - Items listed in the wall-mounted items list MUST use a wall as the reference object and "hang" as the relative position.
   - They cannot be placed on the floor, on other furniture, or in a corner.
</constraint>

<example>
<func> place_object("sleeping area_1", [2500, 1000, 2000], ["east wall"], ["center"]) </func>
<func> place_object("sofa_1", [2200, 800, 950], ["room"], ["southwest corner of"]) </func>
<func> place_object("work_lamp_1", [200, 370, 200], ["desk_1"], ["on"]) </func>
<func> place_object("frame_1", [800, 600, 30], ["pink wall"], ["hang"]) </func>
</example>

<instruction>
Using the suggestions provided in: {suggestions}, do the following for each element (bubble or single furniture):
- Identify the element name.
- Extract or infer the size (use [0, 0, 0] if not available).
- Determine the correct reference object.
- Correctly identify the relative position, applying all constraints and rules.

Format each result as a separate function call using the required syntax. Only return the <func>...</func> blocks, one per item, in the original order. Do NOT generate <func> calls for individual furniture items inside a bubble.

CRITICAL: You MUST produce exactly one <func> block for EACH of the following items (use the EXACT names shown): {required_items_list}
If a suggestion is missing or unclear for an item, infer a reasonable placement from the room context, but you MUST still output a <func> for it. The total number of <func> blocks must equal the number of items listed above.
</instruction>
\end{lstlisting}

\subsection{Wall-Mounting Classification}
\label{supp:mounting}
A preprocessing call that classifies each inventory item as wall-mounted
or floor-standing; wall-mounted items are deferred to the end of the
placement order.
\begin{lstlisting}
<instruction>
Determine for each furniture item in the list whether it should be wall-mounted (hung on / affixed to a wall at elevation) or placed on the floor.

Return one <func> block per item using ONLY this format:
<func> set_mounting("item name", true|false) </func>

Definitions:
- Wall-mounted (true): Items designed to hang or attach vertically (e.g., frame, artwork, painting, poster, mirror, wall shelf, floating shelf, wall cabinet, wall-mounted TV, wall lamp, sconce, coat rack, pegboard, hook rail).
- Floor (false): Freestanding or resting items (sofa, chair, armchair, bed, table, desk, side table, nightstand, dresser, wardrobe, floor lamp, rug, bench, coffee table, ottoman, storage box).

Rules:
1. If the name clearly indicates a wall-hung variant (e.g., "wall shelf", "floating shelf", "wall cabinet", "wall lamp", "sconce"), classify as true.
2. Generic terms ("shelf", "cabinet", "lamp") are false unless explicitly prefixed by wall/floating/sconce.
3. TV:
   - "wall-mounted tv" or "tv (wall)" → true
   - plain "tv" → false
4. Ambiguous or unknown items default to false.
5. Mirrors/frames/art are always true unless specified as "floor mirror".
6. Coat rack / pegboard / hook rail → true.
7. Do not infer mounting from size if not stated.
8. No additional commentary—only the <func> lines.

Input furniture list: {furniture_list}
</instruction>

<example>
<func> set_mounting("single_furniture_sofa_1", false) </func>
<func> set_mounting("single_furniture_wall shelf_2", true) </func>
<func> set_mounting("single_furniture_frame_large", true) </func>
<func> set_mounting("single_furniture_floor lamp_black", false) </func>
<func> set_mounting("single_furniture_mirror", true) </func>
<func> set_mounting("single_furniture_tv", false) </func>
<func> set_mounting("single_furniture_wall-mounted tv", true) </func>
</example>

<instruction>
Output one <func> block per item in the original order. Only the blocks.
</instruction>
\end{lstlisting}

\section{Plausibility Judge Prompt}
\label{supp:judge}

The full prompt of the VLM judge (Gemini~3.1~Pro) used for the
plausibility score, including the complete four-tier grading rubric
summarized in the main paper. The judge receives a rendering of the
generated layout and returns a score in $\{1,2,3,4\}$ with free-text
reasoning in JSON format.
\begin{lstlisting}
The generated scenelayout of the room is shown in the image provided.
Please evaluate the **holistic arrangement of all furniture** in the room on a scale of 1-4.

**Important:** Evaluate the layout as a whole. A room often contains multiple pieces of the same furniture type (e.g., several chairs) serving different functions. Judge each piece in its own functional context — a desk chair facing a wall is correct, a dining chair facing a dining table is correct, and an accent chair against a wall can also be correct. Do NOT penalize the layout just because one piece of a type faces a direction that seems wrong for another piece of the same type.

**Important — surplus items:** The furniture list is **chosen by the user**, so the room may contain **more pieces than fit into tidy functional groups** (extra chairs, an extra table, spare rugs, a stray bench, an extra cabinet, etc.). The generator must place every provided item, so some pieces will inevitably end up lined along walls, in corners, or as leftovers. **This is an inventory constraint, NOT a design failure.** Judge the layout primarily on whether the **main functional groups** (e.g., the dining set, the primary seating area, the bed zone) are coherent. Do NOT lower the score just because a few surplus/extra pieces sit against a wall or don't belong to a clear group — that is expected and acceptable.

**Important — open center is fine:** Furniture pushed toward the walls with an open, empty area in the middle of the room is a **normal, valid arrangement** that supports circulation. An empty center is NOT a void, NOT "awkward", and NOT a defect on its own.

**Grading Rubric:**

**1 = Critical Physical or Semantic Failure:**
- **Floor-standing** furniture (tables, sofas, beds, chairs) is floating in mid-air with no support, or objects are heavily intersecting with *each other* (e.g., a table inside a sofa).
- Furniture extends **clearly and unambiguously** beyond the room boundary (out of room). **The visible floor is the room boundary.** A piece counts as out of room **ONLY** when you can clearly see an **empty gap / void** (background, blank space, or a different surface that is NOT the floor) directly **underneath a large portion** of the piece, so it is unmistakably hovering past where the floor ends. You must actually see floor missing beneath it.
- **Strongly bias toward IN-BOUNDS. Do NOT flag these very common false positives:**
  - **Corner / against-wall placement is correct and expected.** Beds, sofas, and storage are *supposed* to sit flush at the perimeter of the floor, with their base aligned to the floor edge. Touching, aligning with, or sitting right at the floor edge is NOT out of room.
  - **Perspective foreshortening:** in a 3D interior view the far edge of a large piece (e.g. a bed in a corner) can visually appear to cross the floor outline even when it is fully inside. Do not penalize unless you see an actual empty gap under it.
  - **Shadows / dark contact patches** under or beside furniture are NOT part of the object and NOT a void — ignore them entirely.
  - If the floor is still visible (or hidden by the piece itself) right up to and under the piece, it is inside the room.
  - If you are not near-certain there is empty void beneath a large part of the piece, treat it as inside (when in doubt, IN-BOUNDS). Do not give score 1 for boundary reasons based on a piece merely being *close* to the edge.
- **Note:** Wall-mounted items (paintings, shelves, wall lamps, mirrors, clocks, TVs) are expected to be elevated on the wall — this is NOT a physics error.
- **Score 1 is reserved for PHYSICAL / boundary failures** (floating, heavy intersection, clearly out of room). Purely semantic problems (poor grouping, broken pairings, scattered furniture) are **never** a score 1 — the worst a semantically poor but physically valid layout can get is a score 2.

**2 = Structural Failure (a MAIN furniture group is genuinely broken):**
Score 2 only if a **primary** part of the layout is genuinely broken — i.e. **any ONE** of the following:
- A **major piece is not anchored** — a bed, sofa, or large storage unit floats in the open middle of the room instead of being placed against a wall where it belongs. (A piece against a wall, even if it looks plain, is fine.)
- **The single primary functional group does not exist at all** — e.g. a living room where the main seating is genuinely not grouped together AND not oriented toward any focal point, or a dining set where the table has essentially no chairs around it. A *partially* imperfect group (a couple of chairs angled oddly, one armchair off to the side) is NOT enough — that is score 3.
- **Circulation is blocked** — you genuinely could not walk from the entry/edge to the main areas without climbing over or squeezing past furniture.

The objects obey basic physics (on the floor, not heavily overlapping, inside the room), but a MAIN group above is genuinely broken.
- **Do NOT reach score 2 by accumulating minor issues.** Surplus/extra items along the walls, a spare rug, an odd bench, a few chairs facing slightly off, or an empty center are NOT structural failures — these stay at score 3.

**3 = Functional and Coherent (DEFAULT — use this whenever the main groups work):**
This is the default score whenever the **main functional groups are recognizable and broadly sensible**, and none of the score-2 failures apply.
- The primary group(s) read as groups (the dining set is together, the main seating reads as a seating area), major pieces are anchored, and circulation is passable.
- **Minor flaws are fully allowed and expected here** — surplus/extra items parked against walls, a stray rug or bench, a few chairs angled imperfectly, slight clutter, generic styling, an empty center, or uneven spacing. Several such minor imperfections together still stay at 3 as long as the main groups function.
- It simply lacks a standout, polished, designer-quality touch (otherwise it would be a 4).

**4 = Excellent / Professional Design (polished main groups + a design highlight):**
Score 4 **only if BOTH** conditions hold:
1.  **The main functional groups are polished** — cleanly oriented, well aligned, and good contact, with no awkward orientations or clutter *within the main groups*. **Surplus/extra items neatly parked against walls or in corners do NOT disqualify a 4** — they are an unavoidable inventory constraint, not a defect, as long as they are tidy and out of the way. AND
2.  **At least one clear design highlight**, such as:
    - a balanced / symmetric composition,
    - generous negative space with comfortable, inviting circulation (not merely passable),
    - tightly grouped, naturally aligned furniture relationships,
    - or a clear visual focal point (e.g., a seating group composed around a TV / fireplace as a conversation center).

If the main groups are clean but plain (no highlight), cap it at 3. A score of 4 must be earned by a visible, describable design strength — keep it reserved for genuinely impressive layouts, not the default.

**Instructions for Response:**
1.  **Analyze the Scene:** Briefly describe the room type and identify the main furniture groups and their intended functions.
2.  **Check Physics & Boundary:** Are there floating objects or major collisions between objects? For the room boundary, only flag a piece as out of room when you can clearly see an **empty void underneath a large portion of it** (floor visibly missing beneath it). A bed or sofa placed flush in a **corner or against a wall** is correct — sitting at the floor edge is expected, NOT out of room. Ignore shadows and perspective foreshortening, and when not near-certain, treat the piece as in-bounds. Only a clear, obvious void beneath the piece forces a score of 1.
3.  **Check Main Groups:** Identify the **main functional group(s)** (dining set, primary seating area, bed zone) and judge only whether *those* read as coherent groups — roughly together and sensibly oriented (e.g., chairs around the table, coffee table near the sofa, nightstand by the bed). Do NOT penalize **surplus/extra items** parked against walls or in corners, and remember different pieces of the same type may serve different roles. A main group only fails (score 2) if it genuinely does not exist as a group at all — minor imperfections stay at 3.
4.  **Check Connectivity (visual):** Is there a passable open path to move through the room? An **empty/open center with furniture along the walls is fine** and counts as good circulation. Only treat connectivity as a problem (score 2) if furniture genuinely blocks movement across the room.
5.  **Final Score:** Output the score in JSON format. A layout that is functional and makes sense overall should score at least 3. **Tie-break rule:** if you are torn between two scores, pick the HIGHER one unless you can name a concrete, specific defect that justifies the lower score.

Your output must be strictly in the following JSON format:
```json
{
"reasoning": "YOUR DETAILED REASONING HERE",
"final_answer": 3
}
```
\end{lstlisting}

\end{document}

%% file: sections/abstract.tex
\begin{abstract}
  Generating physically plausible 3D room layouts is essential for home furnishing retail, enabling customers to visualize products in their own homes and confidently make purchasing decisions. However, a gap exists between academic research and real-world application: existing solutions primarily focus on algorithmic strategies for furniture placement, largely neglecting the non-rectangular geometries and strict door/window constraints prevalent in real homes. To bridge the gap, we introduce a hybrid, hierarchical framework tailored for retail, specifically designed to support scalable spatial planning applications. Our system decouples generation into three stages: (1) functional furniture clustering and fine-grained intra-zone placement; (2) macro-routing guided by a vision-language model (VLM) to anchor both these clustered zones and any remaining standalone furniture within diverse polygonal boundaries; and (3) rule-based optimization for collision-free micro-arrangements that respect architectural constraints. We evaluate our system on production-scale catalogs and a representative set of irregular real-world topologies. Our results show that our approach attains the highest perceptual plausibility while maintaining good geometric compliance at relatively low latency, and extends to irregular boundaries that existing methods do not natively support.
  \keywords{Spatial Planning \and Hybrid Layout Generation \and VLM}
\end{abstract}

%% file: sections/introduction.tex
\section{Introduction}
\label{sec:intro}

\begin{figure*}[t]
  \centering
  \includegraphics[width=\linewidth]{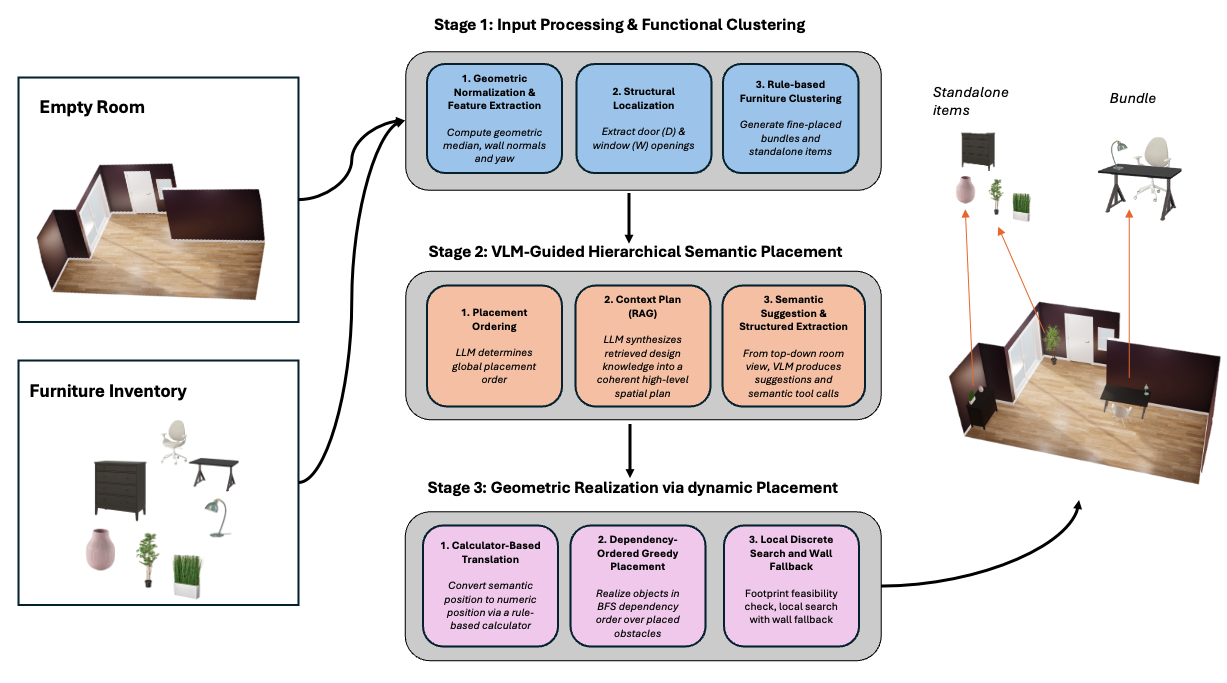}
    \imgcopyright
    \caption{\textbf{The \textsc{PolyLayout} pipeline.} Stage~1 parses the room
    topology and groups the inventory into functional bundles with rule-based
    internal arrangements; Stage~2 (vision-language model, VLM) positions each bundle and standalone
    item over an annotated top-down rendering of the polygonal floor plan;
    Stage~3 resolves the proposals into precise, collision-free 3D poses in
    dependency order.}
  \label{fig:workflow}
\end{figure*}

Arranging furniture into functional, aesthetically coherent 3D room layouts is increasingly central to home furnishing retail, where digital planning tools support customers both online and in store. In practice, customers furnishing a space face a blank-room problem: presented with an empty floor plan, many struggle to take the first step, and abandon the planning process before engaging with the product catalog. An automatically generated initial layout lowers this barrier, offering a concrete, plausible starting point that customers can react to and refine. The core difficulty, however, is twofold: a valid layout must simultaneously respect the functional relationships between items (\eg, aligning a TV with a sofa) and the hard architectural constraints of the space (\eg, doors, windows, and non-rectangular boundaries).

Traditional layout generation methods have progressed through convolutional neural network (CNN)-based placement \cite{wang2018deep}, transformer architectures \cite{wang2021sceneformer}, and diffusion models \cite{tang2024diffuscene}. However, these data-driven paradigms treat layout generation as a black-box mapping with no mechanism to enforce hard spatial constraints: small per-object errors compound into collisions or violated ergonomic relationships, and nothing guarantees that a predefined inventory is placed logically within a specific room topology.

Recent work has explored Large Language Models (LLMs) for layout generation, leveraging their spatial reasoning and natural language understanding \cite{feng2023layoutgpt, yang2024holodeck}, and more recent efforts extend this to Vision-Language Models (VLMs) that incorporate visual context \cite{sun2025layoutvlm}. Although such models excel at semantic decisions, identifying functional zones and their relative positions, they lack geometric precision: an LLM might suggest placing a desk ``against the wall,'' but cannot compute exact coordinates that account for drawer clearance, chair positioning, and accessibility. Moreover, these systems confine placement to rectangular room specifications: even when doors and windows are modeled \cite{yang2024holodeck}, the irregular polygonal boundaries ubiquitous in real homes are not treated as hard constraints. In a retail setting, however, an automated layout is unacceptable if a selected wardrobe blocks a doorway or protrudes beyond an alcove. Beyond geometric feasibility, real product catalogs carry product-specific placement rules (\eg, tall storage units must be placed against a wall to meet anchoring and tip-over safety requirements) that semantic reasoning alone cannot guarantee.

To bridge this gap, we introduce a hybrid, hierarchical framework tailored for retail (\cref{fig:workflow}). Given a room topology and a predefined furniture inventory, our system decouples generation into rule-based functional
clustering, VLM-driven macro-routing over the polygonal floor plan, and deterministic geometric constraint resolution.

Our contributions are as follows:
\begin{itemize}
    \item \textbf{A hybrid hierarchical pipeline:} An end-to-end architecture that decouples layout generation into clustering, macro-routing, and micro-refinement.
    \item \textbf{Constraint-aware macro-routing:} A VLM-guided positioning strategy that translates irregular polygonal boundaries and fixed doors and windows into actionable spatial anchors for real-home floor plans.
    \item \textbf{Industry-ready micro-arrangement:} Deterministic, rule-based geometric algorithms for final constraint resolution, with the collision-aware guarantees required for retail use.
\end{itemize}

%% file: sections/related_work.tex
\section{Related Work}
\paragraph{Layout Generation with Deep Learning.}
Early data-driven approaches established foundations for indoor scene synthesis by learning spatial relationships from annotated datasets, including example-based synthesis of 3D object arrangements~\cite{fisher2012example}, probabilistic models~\cite{merrell2010computer, yu2015clutterpalette} and graph-based representations~\cite{qi2018human, zhang2021fast}. The introduction of deep generative models significantly advanced the field: DeepSynth \cite{wang2018deep} pioneered CNN-based sequential placement, while FastSynth \cite{ritchie2019fast} factorized the task into separate category, location, orientation, and dimension prediction. Transformer-based methods like ATISS \cite{paschalidou2021atiss} and SceneFormer \cite{wang2021sceneformer} enabled autoregressive generation of object parameters, while diffusion-based approaches including DiffuScene \cite{tang2024diffuscene} and DeBaRA \cite{maillard2024debara} further improved generation fidelity.

\paragraph{LLM- and VLM-based Scene Generation.}
The emergence of Large Language Models has introduced new paradigms for controllable scene synthesis through natural language. LayoutGPT \cite{feng2023layoutgpt} was among the first to prompt LLMs to generate structured object parameters from text descriptions, and subsequent work has explored scene graph generation \cite{ccelen2024design, deng2025global, gao2024graphdreamer} and the direct control of procedural systems \cite{sun20253d, liu2024controllable, hu2024scenecraft}. Holodeck \cite{yang2024holodeck} supports multi-round conversational generation that couples object selection with placement, and CityCraft \cite{deng2024citycraft} showed LLMs' effectiveness in large-scale urban planning. More recently, vision-language models (VLMs) have been applied to layout generation: LayoutVLM~\cite{sun2025layoutvlm} grounds placement in rendered visual
context, improving spatial awareness over text-only LLMs. However, these approaches are
typically demonstrated on simplified rectangular rooms, lack a mechanism to
encode irregular polygonal boundaries and fixed openings as hard constraints, and fall short of the precision needed for detailed furniture arrangement.

\paragraph{Rule-based and Hybrid Approaches.}
Classical optimization-based methods~\cite{yu2011make, merrell2011interactive,
kan2018automatic} formulate layout as cost-function minimization, producing arrangements
that respect functional relationships and spatial guidelines. More recently,
ProcTHOR~\cite{deitke2022} and Infinigen Indoors~\cite{raistrick2024infinigen} combine
procedural constraints with data-driven generation to promote physical plausibility. Our work builds on this constraint-driven direction but targets the retail setting: rule-based algorithms resolve each functional bundle's internal
geometry, while a VLM positions the bundles and remaining standalone items globally over the room polygon, accounting for non-rectangular boundaries and fixed openings.

%% file: sections/methods.tex
\section{Methods}

We formulate automated spatial planning as a decoupled, hierarchical process rather than an end-to-end black-box optimization. \Cref{sec:pipeline} details the core pipeline and \cref{sec:fine_placement}
the fine-grained algorithms that arrange the items within each bundle.

\subsection{Pipeline Architecture}\label{sec:pipeline}

Our pipeline comprises three stages (\cref{fig:workflow}): input processing and rule-based functional clustering (Stage~1), VLM-guided semantic placement (Stage~2), and geometric realization via dynamic placement (Stage~3).

\subsubsection{Stage 1: Input Processing and Functional Clustering}

The system accepts two inputs: room geometry in glTF format and a predefined furniture inventory of the items to be placed.

\paragraph{Room Feature Extraction.}
We distill the raw glTF scene graph into an actionable spatial representation: floor polygons, wall planes, and the vertex coordinates of structural openings. For \emph{geometric normalization}, rather than approximating the room with an axis-aligned bounding box, we reconstruct the true floor polygon by traversing the bottom endpoints of the walls in clockwise order, which preserves non-rectangular (\eg T-shaped) plans exactly. Let $\{\mathbf{v}_i\}_{i=1}^{N}\subset\mathbb{R}^{2}$ denote the ordered floor vertices on the $XZ$ ground plane. We define the room's \emph{reference center} as the geometric median of these vertices (computed via Weiszfeld iterations), which, unlike the arithmetic centroid, yields a stable interior reference on elongated plans. Each wall $w$ is assigned an inward-facing unit normal $\mathbf{n}_w$ derived from the clockwise edge orientation, and a continuous yaw $\theta_w=\operatorname{atan2}(d_z,d_x)$ computed from its $A\!\to\!B$ direction $\mathbf{d}=(d_x,d_z)$, so that an object's local $+X$ axis aligns with a wall at any orientation. For \emph{structural element localization}, each door and window is associated with its parent wall via metadata, yielding per-wall opening sets $\mathcal{D}$ (doors) and $\mathcal{W}$ (windows).

\paragraph{Furniture Clustering.}
To reduce combinatorial complexity, the inventory is partitioned by a rule-based, category-driven procedure into \emph{functional bundles}---coherent groups requiring coordinated placement (\eg, a bed with nightstands)---and \emph{standalone items}. Concretely, each catalog item carries a product-type label, and a fixed
taxonomy maps $212$ product types onto $43$ functional roles across the six
area types of \cref{sec:fine_placement}. A bundle is instantiated only when
its mandatory anchor role is present (\eg, a bed for a sleep area), with support items attached up to per-role caps. Product types admissible in several areas (\eg, rugs)
are resolved by a fixed area-priority order, each instantiated bundle
consuming its items before the next area type is considered; items matching
no role remain standalone. Within each bundle, a fine-placement step (\cref{sec:fine_placement}) arranges the constituent items relative to one another using rule-based templates (\eg, positioning nightstands flush on either side of the bed), yielding a standard local configuration. Each functional bundle $B_j$ is thereafter treated as a composite object with an aggregate bounding box $(w_j,d_j,h_j)$ and a designated anchor.
 
\subsubsection{Stage 2: VLM-Guided Semantic Placement}

For each functional bundle and standalone item, placement follows a semantic reasoning pipeline.

\paragraph{Placement Ordering.} The VLM first determines a global placement order over bundles and standalone items, ensuring that large anchoring structures are positioned before items that depend on them. Wall-mounted items are deferred to the end of the order.

\paragraph{Context Plan via Retrieval-Augmented Generation.} A high-level spatial plan is constructed via Retrieval-Augmented Generation (RAG) over curated interior-design content (furnishing guidelines,
room-specific heuristics, and activity-based spatial requirements): the room context is encoded into a structured query, and the VLM synthesizes the retrieved passages into a plan specifying functional-zone allocation and
inter-item relationships, which is injected into the placement prompt.

\paragraph{Semantic Suggestion.} The VLM receives a multi-modal context: (i) a top-down rendering of the room in which wall segments carry distinct chromatic labels and structural openings are overlaid as colored markers (doors in red, windows in blue); (ii) a structured textual description containing room dimensions, per-wall door/window counts, and the wall ordering; and (iii) the dimensions of each bundle and item. To keep the output feasible, each placement selects a single \texttt{relative\_position} from a reference-dependent set: wall segments admit \{``center'', ``hang''\}; the room admits its pairwise-adjacent corners, named by their two adjoining wall labels (\eg, the ``red--green corner''); windows admit \{``under''\}, doors \{``next to''\}, and
previously placed objects \{``in front'', ``left side'', ``right side'', ``on'', ``under''\}, with left/right defined by the wall-orientation convention fixed in Stage~1. The full prompt template is provided in the supplementary material.

\paragraph{Structured Extraction.} The semantic suggestions are parsed into executable placement functions:
\begin{verbatim}
<func>place_object(placement_id, [length, height, width],
                   reference_object, relative_position)</func>
\end{verbatim}
Extraction enforces strict geometric admissibility: bundles are restricted to room corners or wall centers to prevent spatial fragmentation. An overlap-prevention rule disqualifies any wall segment containing a structural opening from ``center'' placement, forcing the agent to use a ``next to'' descriptor and preserve functional clearance.

\subsubsection{Stage 3: Geometric Realization via Dynamic Placement}

The semantic suggestions are realized as precise, collision-aware 3D poses by a deterministic \emph{dynamic placement} solver. Because each item carries an explicit structural reference, the mapping from a \texttt{(relative\_position, reference\_object)} pair to a metric pose is well-defined; the solver's role is to locally resolve residual infeasibilities, out-of-bounds protrusion, opening obstruction, and inter-object overlap.

\paragraph{Calculator-Based Translation.}
An analytic calculator maps each structured call to a target pose $(\mathbf{t}^{*}_j,\theta_j)$, with anchor $\mathbf{t}^{*}_j=(x_j,z_j)$ and orientation $\theta_j\in[0,2\pi)$, using the clockwise wall ordering and inward normals from Stage~1. For example: ``center of [wall]'' maps to the wall midpoint oriented perpendicular to the wall, and ``next to [door]'' to an adjacent pose preserving swing clearance.

\paragraph{Dependency-Ordered Greedy Placement.}
Objects are placed one at a time, in a dependency order given by a breadth-first
traversal of the \texttt{reference\_object} graph; objects already placed act as
hard obstacles for the current one. A candidate pose is accepted only if it is
\emph{feasible}: the object must lie fully inside the room, keep every door and
window clear, and overlap no placed object. Writing
$\mathrm{fp}(o)\subset\mathbb{R}^{2}$ for the $XZ$ footprint of object $o$,
$\mathcal{F}$ for the floor polygon, and $\mathcal{O}_{\mathrm{placed}}$ for the set of placed
objects, this feasibility test is
\begin{equation}
\Phi(o)\;=\;\big[\,\mathrm{fp}(o)\subseteq\mathcal{F}\,\big]\ \wedge\ \big[\,o \text{ obstructs no opening}\,\big]\ \wedge\ \big[\,A_{\mathrm{ovl}}(o)=0\,\big],
\label{eq:feasibility}
\end{equation}
where the total overlap area against placed objects is
\begin{equation}
A_{\mathrm{ovl}}(o)\;=\;\sum_{o'\in\mathcal{O}_{\mathrm{placed}}}\operatorname{Area}\!\big(\mathrm{fp}(o)\cap\mathrm{fp}(o')\big).
\label{eq:overlap}
\end{equation}

\paragraph{Local Discrete Search and Wall Fallback.}
If the calculator target $\mathbf{t}^{*}$ already satisfies $\Phi$, it is
accepted directly. Otherwise, the solver searches a discrete set of offsets
around the semantic anchor. Let $\mathcal{A}(o)$ denote the search directions
of object $o$: the wall-parallel unit axis for wall-mounted items, plus the
inward wall normal for free-standing items. The candidate poses are
\begin{equation}
\mathbf{t}^{(k,\hat{\mathbf{d}})}
  =\mathbf{t}^{*}+k\,\Delta\,\hat{\mathbf{d}},
\qquad \hat{\mathbf{d}}\in\mathcal{A}(o),\quad
k\in\{\pm1,\dots,\pm K\},\quad
K=\Big\lfloor \tfrac{W_{\mathrm{wall}}}{\Delta}\Big\rfloor,
\label{eq:step}
\end{equation}
with step $\Delta=100$\,mm and the deviation bounded by the reference wall
width $W_{\mathrm{wall}}$ (a fixed $3000$\,mm cap is used for free-standing
items lacking a wall reference). Candidates are evaluated in order of
increasing $|k|$, and the first feasible pose is returned. If no zero-overlap
pose exists on the original reference wall, a wall-stacked item is re-anchored
to nearby walls and the search is repeated. When every candidate still
overlaps, the item is either assigned the globally minimum-overlap pose
\begin{equation}
\mathbf{t}^{\star}
  =\arg\min_{k,\hat{\mathbf{d}}}\,
   A_{\mathrm{ovl}}\!\big(o(\mathbf{t}^{(k,\hat{\mathbf{d}})})\big),
\label{eq:minoverlap}
\end{equation}
or reported as \emph{unplaced} rather than forced into a hard collision.

%% file: sections/fine_placement.tex
\subsection{Fine-Grained Placement}\label{sec:fine_placement}

The fine placement module deterministically resolves the intra-bundle geometry that semantic models cannot reliably specify: exact
coordinates, clearances, and surface arrangements within each functional bundle.

\subsubsection{Functional Area Processing}
We organize intra-bundle placement into six functional area types, each governed by domain-specific algorithms and anchored by a mandatory item:work areas (desk), sleep areas (bed), dining areas (table), social areas (sofa), reading areas (seating), and storage areas (bookcase).

\subsubsection{Rule-Based Placement Architecture}

\begin{figure}[t]
  \centering
  \includegraphics[width=0.45\linewidth]{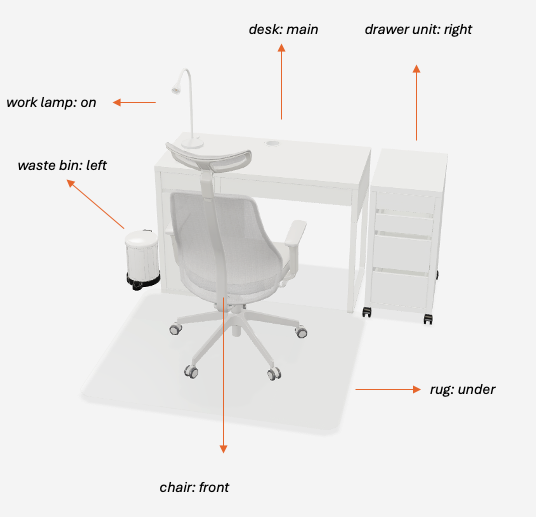}
  \imgcopyright
  \caption{Intra-bundle relative positioning for a work area, with the desk
    as the anchor.}
  \label{fig:fine_placement}
\end{figure}

\paragraph{Relative Positioning.}
Items are positioned relative to an anchor through a relative-position template that encodes each area's standard configuration. In a work area, for instance, the desk serves as the anchor and surrounding items follow the template (\cref{fig:fine_placement}). These relationships are then translated into 3D coordinates $(x,y,z)$ and rotations $(\theta_x,\theta_y,\theta_z)$, accounting for minimum clearances, accessibility, and ergonomic guidelines.

\paragraph{Surface Placement Geometry.}
For placement on surfaces (\eg, desk tops), we (i) extract surface polygons from product
metadata, (ii) compute the unoccupied region via polygon difference against overlapping
elements, and (iii) take the axis-aligned bounding rectangle of the largest unoccupied region as
the usable area, ensuring accessories are placed only on valid, stable surfaces.

\subsubsection{Integration and Efficiency}
Area types are processed iteratively: after one type is placed (\eg, all viable sleep areas), the remaining items are evaluated for the next, maximizing inventory utilization while
preserving coherent groupings. Being deterministic and free of model inference,
the arrangement completes a functional area in well under a second on CPU. The templates are, however, hand-authored: extending beyond our six area types
(\eg, kitchens with appliance-specific clearances) requires new domain rules, which we view as the main scalability cost of the hybrid design.

%% file: sections/experiments.tex
\section{Experiments}
\label{sec:experiments}

We evaluate \textsc{PolyLayout} against two layout generators, \textsc{Holodeck}~\cite{yang2024holodeck} and \textsc{LayoutVLM}~\cite{sun2025layoutvlm}, on a retail benchmark of real IKEA home furnishing products in real floor plans. All three share the Gemini~2.5~Flash backbone and matched hosted configurations, isolating the placement methodology.

The three methods differ in the room geometries they natively accept. \textsc{LayoutVLM} optimizes object poses within a rectangular extent and provides no mechanism to encode openings or non-convex boundaries as hard constraints. \textsc{Holodeck} does model doors and windows, but as part of its own generation: its LLM designs the floor plan and installs the doorways and windows itself~\cite{yang2024holodeck}, rather than conditioning on an externally specified room with fixed opening positions; its solver likewise
assumes rectangular boundaries. We therefore structure the evaluation in two parts: a controlled three-way comparison on rectangular rooms without openings, the largest setting in which all three methods are natively
applicable (\cref{sec:main-results}), followed by a capability study of \textsc{PolyLayout} on openings and non-rectangular plans (\cref{sec:generalization}).

\subsection{Dataset}
\label{sec:dataset}

\paragraph{Rooms.}
Only the room geometry varies across settings. We use three settings: $3$ rectangular rooms without openings, used for the main comparison; $3$ rectangular rooms with doors and windows; and $5$ non-rectangular plans (L-, T-, U-shaped, beveled, and cut), also with doors and windows. The latter two probe generalization to real-home layouts (\cref{sec:generalization}).

\paragraph{Inventories.}
The same $44$ inventories are reused across all settings. For each room function, Gemini provides the product types; real catalog SKUs are then selected manually for those types, each with metric dimensions and 3D geometry asset: $15$ living rooms (\eg, sofas, armchairs, coffee/side tables, storage, rugs, lamps \etc), $14$ bedrooms (\eg, bed frames, nightstands, wardrobes, dressers, desks, seating \etc), and $15$ dining rooms (\eg, dining tables, chairs, sideboards, lighting \etc), plus decor such as plants and wall art. Each inventory holds $9$--$33$ items (mean $17.1$), and each (room, inventory) pair is evaluated with two independent runs, giving $44\times3\times2=264$ conditions per rectangular setting and $44\times5\times2=440$ for the non-rectangular one.

\subsection{Setup}
\label{sec:setup}

\paragraph{Serving.}
\textsc{PolyLayout} and \textsc{Holodeck} run on CPU; \textsc{LayoutVLM} requires
a GPU ($1\times$L4), as one optimization takes minutes. Latency is end-to-end.

\paragraph{Metrics.}
We measure layouts along three axes. Completeness: \textbf{success rate} (Succ.), the
fraction of conditions yielding a non-empty layout, and \textbf{placement ratio} (Place.),
placed over total items. Physical validity: \textbf{in-bounds} (In-b.), the exact
fraction of item footprints contained in the room polygon. Perceptual
quality: \textbf{plausibility score} (Plaus.), a $4$-tier score (\cref{tab:rubric}) from a VLM
judge (Gemini~3.1~Pro, a different model from the generation backbone;
since both are Gemini models, a same-family bias cannot be fully excluded,
and the human study in \cref{sec:human-eval} serves as a model-free check). The full judge prompt is provided in the supplementary material. We
also report \textbf{Latency} (lower is better), the end-to-end generation time.

\paragraph{Protocol.}
We deliberately avoid automated bounding-box overlap scores, which misclassify
intentional tuck-under (chairs under tables, rugs) as collisions; gross physical
failures are instead reflected in In-bounds and penalised by the plausibility
judge. \textsc{Holodeck} returns an empty room in $9/264$ conditions, which
cannot be perceptually scored; In-bounds and plausibility therefore use the
$N{=}255$ conditions non-empty for all methods, while Success and Placement use
the full $264$.

\begin{table}[t]
  \centering
  \caption{Plausibility rubric used by the VLM judge, from $1$ (critical failure) to $4$ (polished).}
  \label{tab:rubric}
  \begin{tabular}{@{}cl@{}}
    \toprule
    Level & Criterion \\
    \midrule
    1 & Critical physical/boundary failure (floating, overlap, out-of-bounds) \\
    2 & A main functional group is broken \\
    3 & Coherent and functional; minor flaws tolerated (default) \\
    4 & Polished, with a clear design highlight \\
    \bottomrule
  \end{tabular}
\end{table}

\subsection{Main Results}
\label{sec:main-results}

\paragraph{The completeness--validity trade-off.}
On rectangular rooms without openings (\cref{tab:main}), success and in-bounds are near-saturated
for every method (Succ.\ $\geq\!96.6\%$, In-b.\ $\geq\!0.996$), so the two
discriminating axes are perceptual \emph{plausibility} and placement
\emph{completeness}. The baselines sit at opposite ends of this trade-off.
\textsc{LayoutVLM} maximises adherence, achieving the best placement ratio
($0.989$) by packing in nearly every object, but pays for it with the lowest
plausibility ($2.49$): objects are placed but often collide or ignore functional
relations. \textsc{Holodeck} takes the opposite stance, avoiding hard overlaps
through discrete constraint satisfaction; this yields cleaner but conservative
scenes that drop items and, when no feasible assignment exists, return no layout
at all, driving its placement ratio down to $0.827$ with the highest variance
($\sigma\!=\!0.24$). 

\begin{figure*}[t]
  \centering
  \setlength{\tabcolsep}{1pt}
  \begin{tabular}{cccc}
    & \small Bedroom & \small Living Room & \small Dining Room \\
    \rotatebox{90}{\small\textsc{Holodeck}} &
      \includegraphics[width=.30\linewidth]{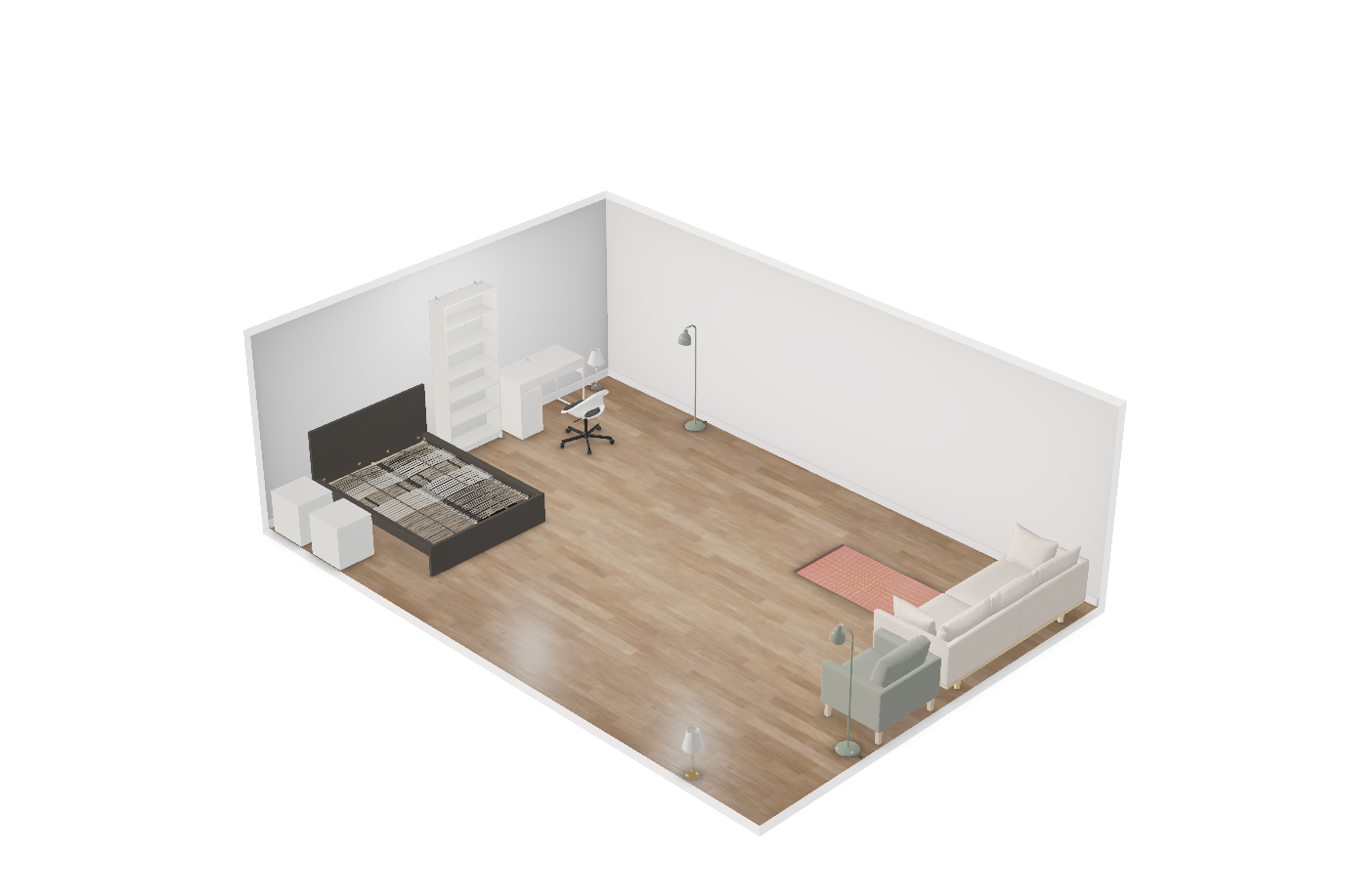} &
      \includegraphics[width=.30\linewidth]{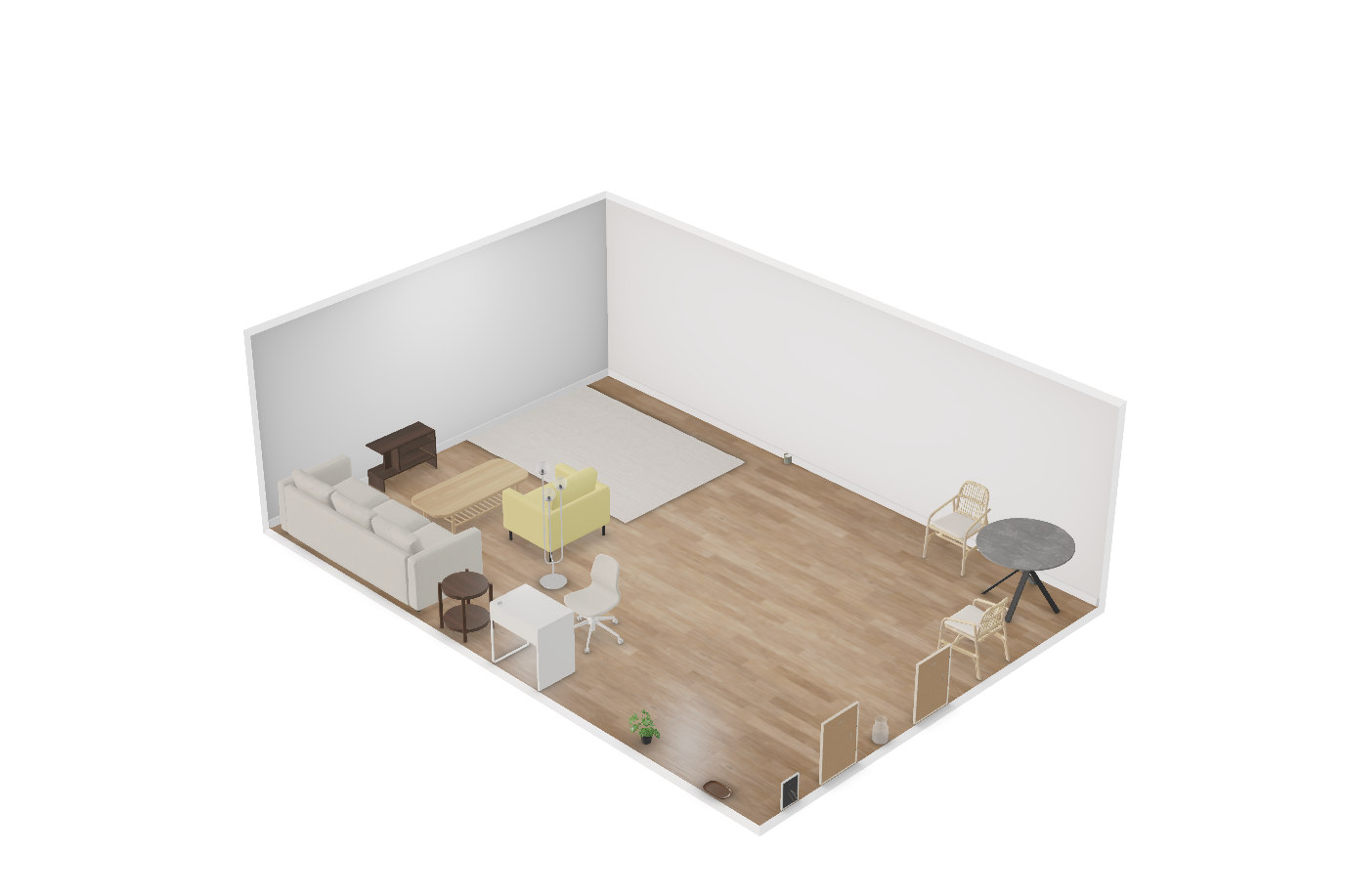} &
      \includegraphics[width=.30\linewidth]{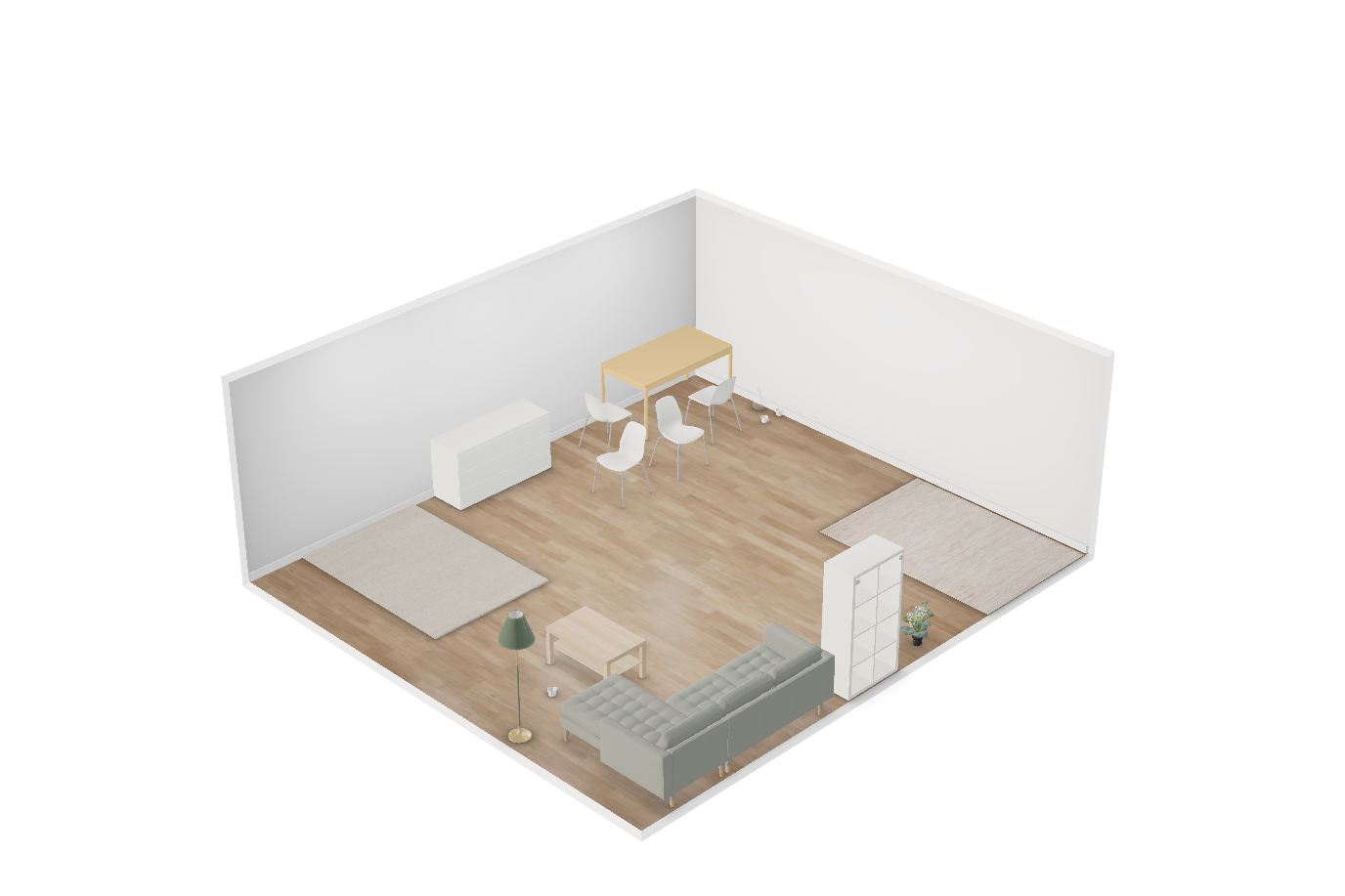} \\
    \rotatebox{90}{\small\textsc{LayoutVLM}} &
      \includegraphics[width=.30\linewidth]{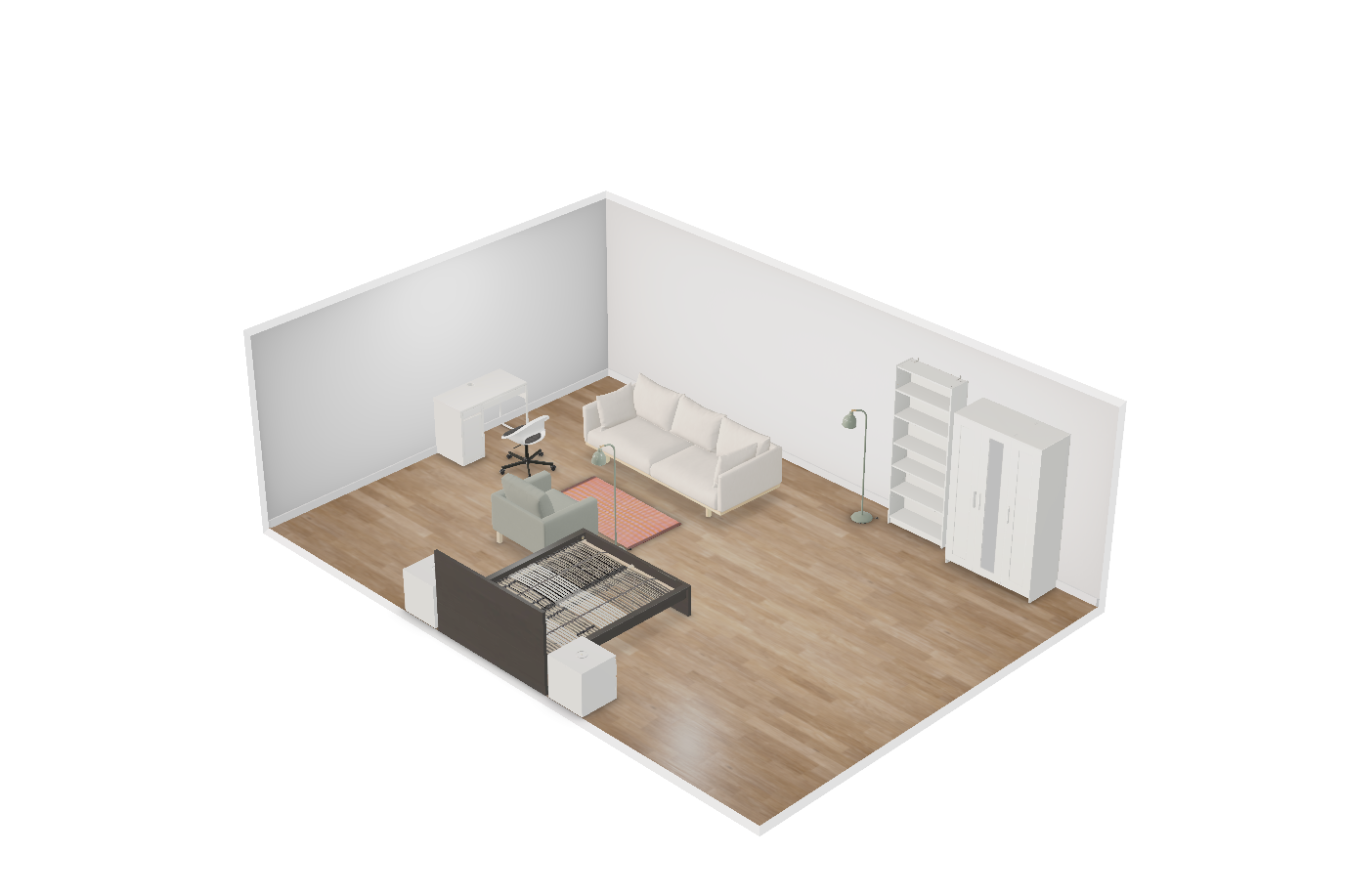} &
      \includegraphics[width=.30\linewidth]{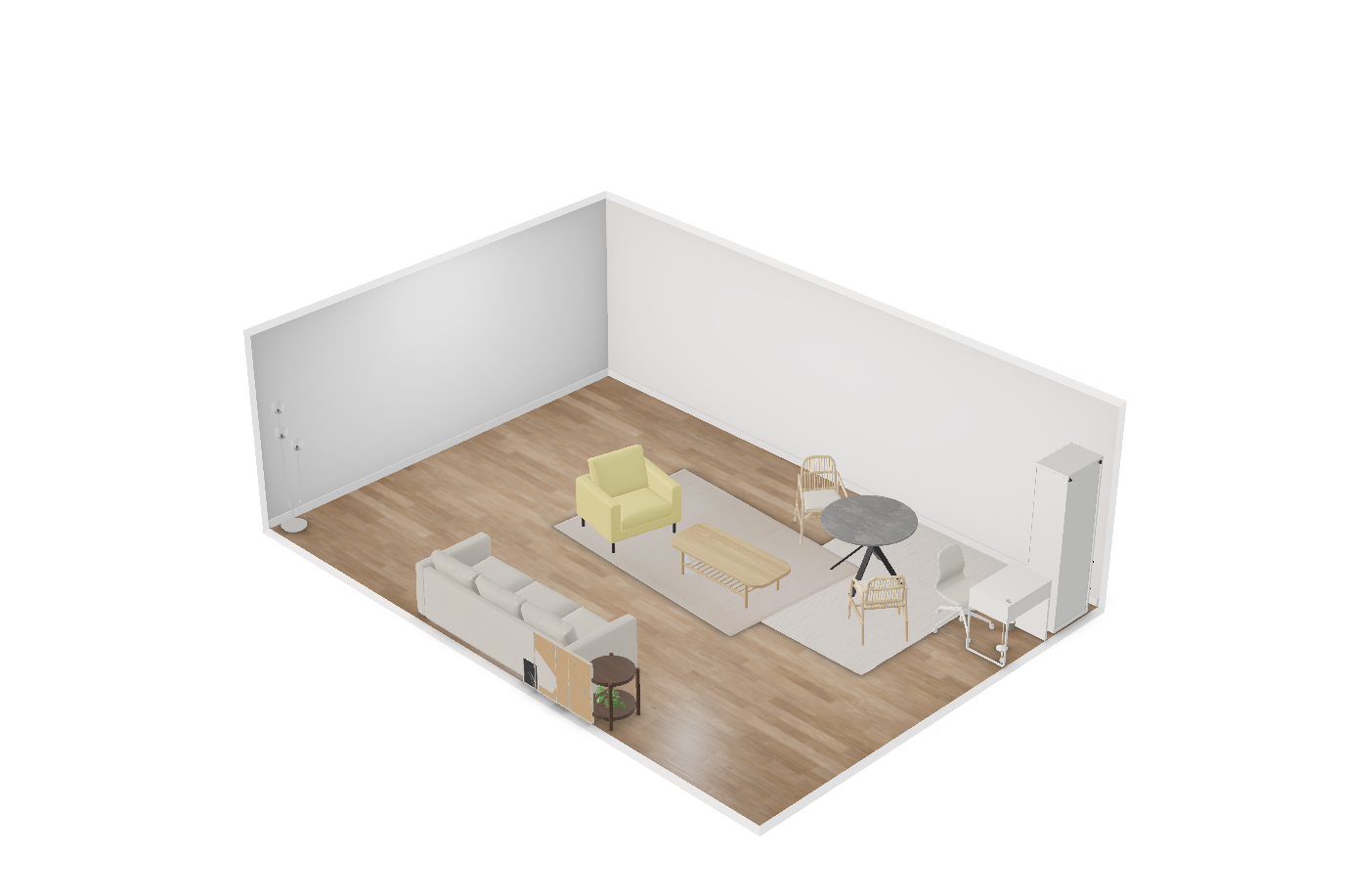} &
      \includegraphics[width=.30\linewidth]{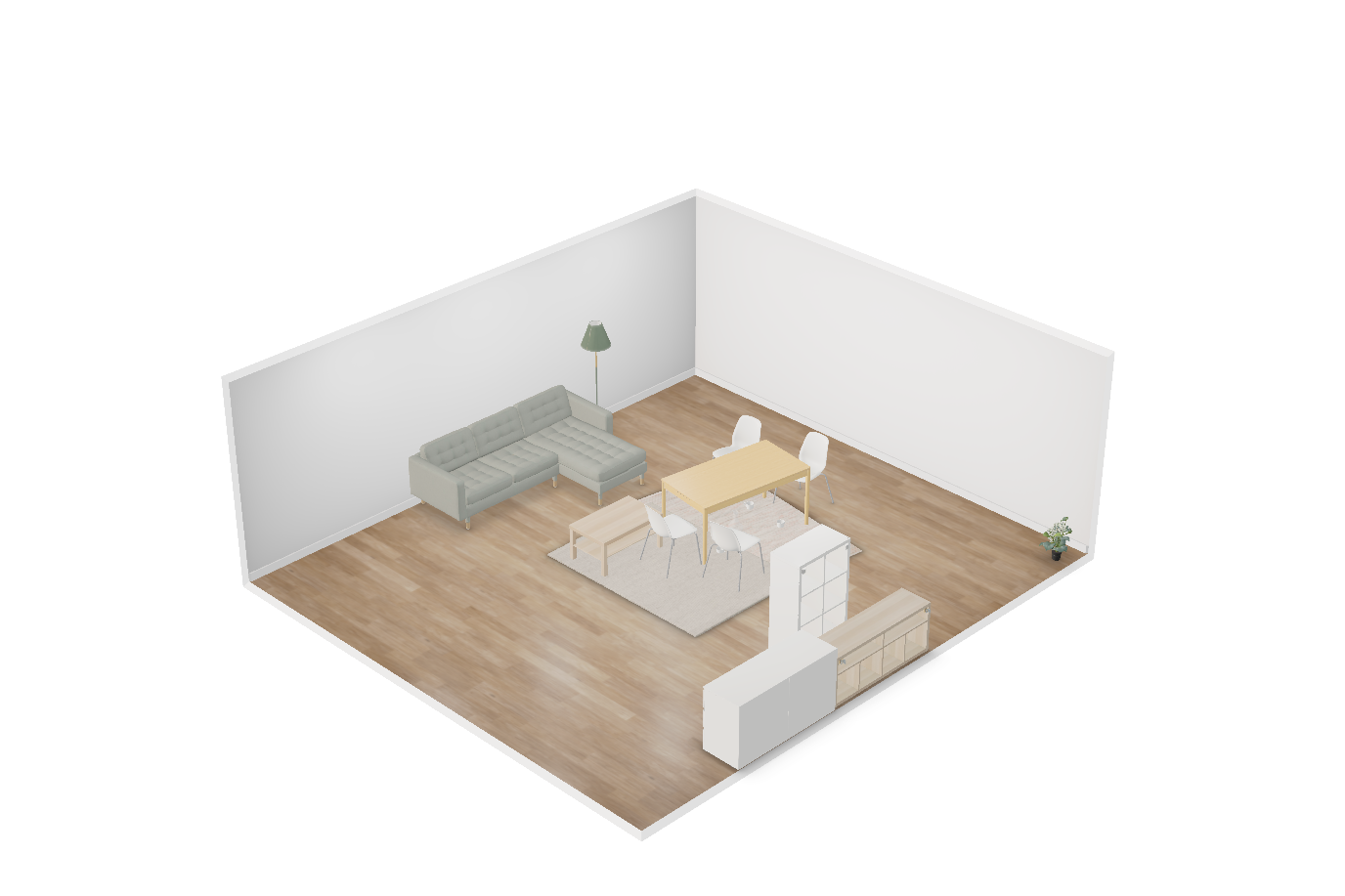} \\
    \rotatebox{90}{\small\textbf{\textsc{PolyLayout}}} &
      \includegraphics[width=.30\linewidth]{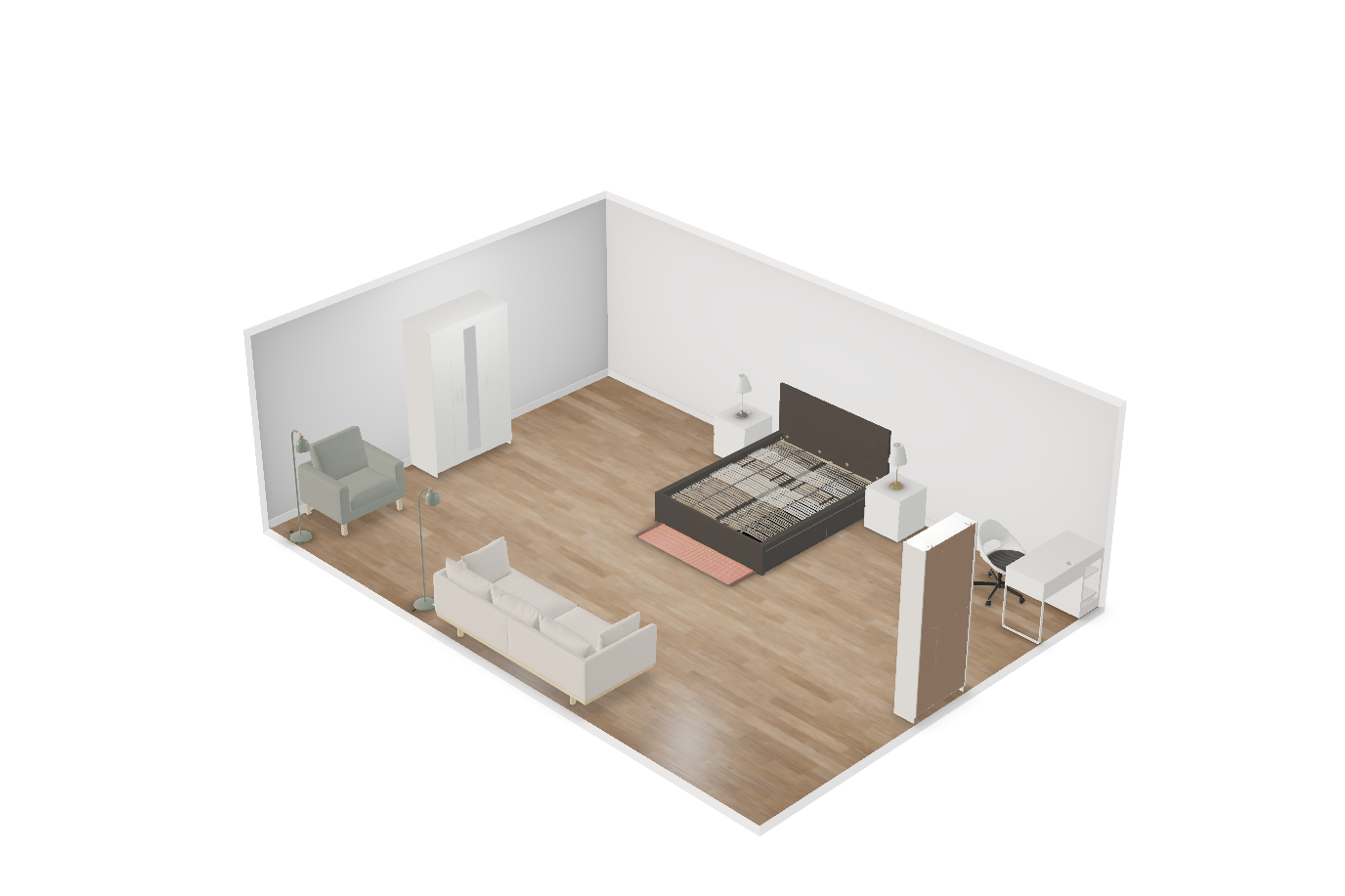} &
      \includegraphics[width=.30\linewidth]{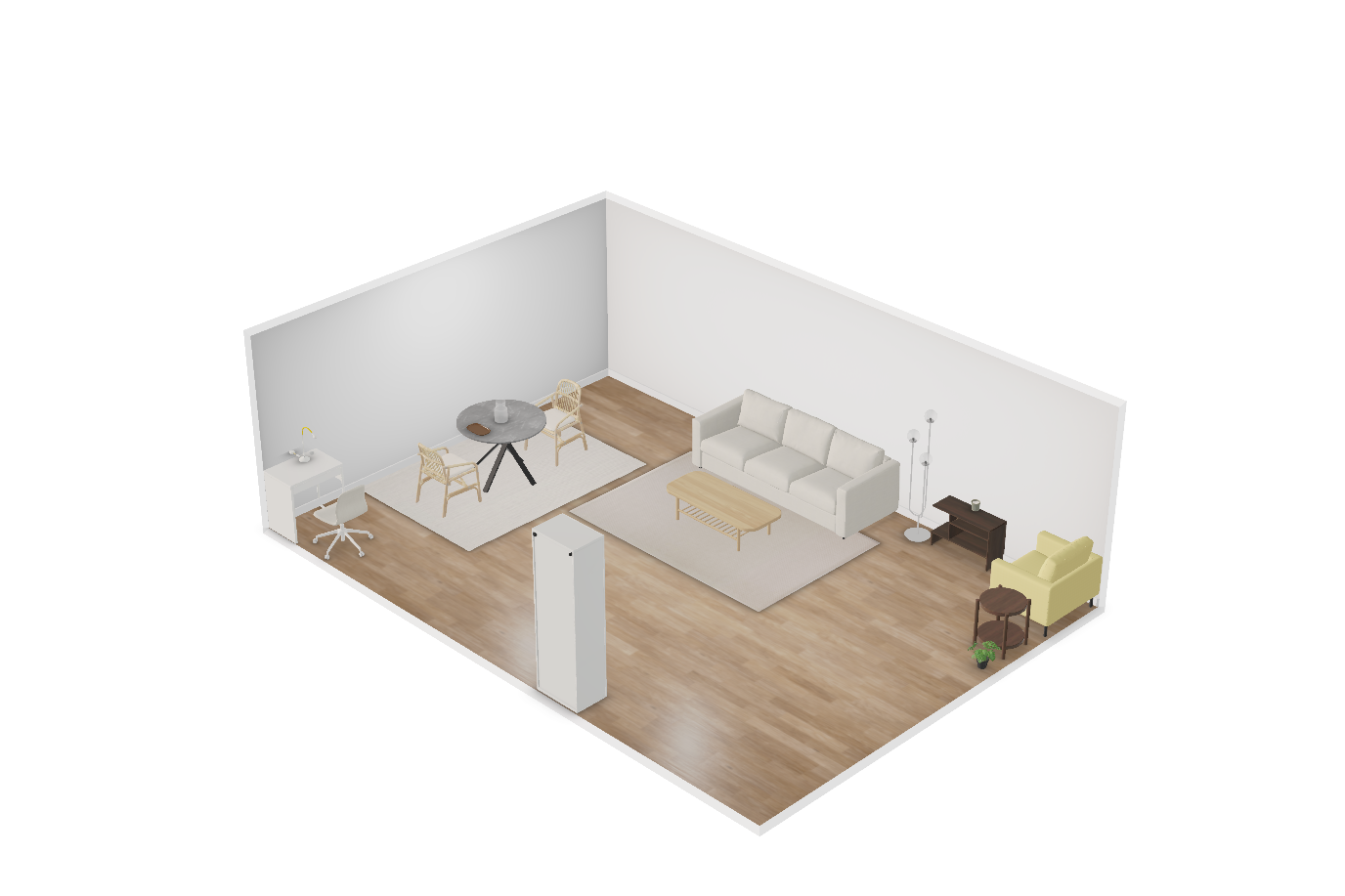} &
      \includegraphics[width=.30\linewidth]{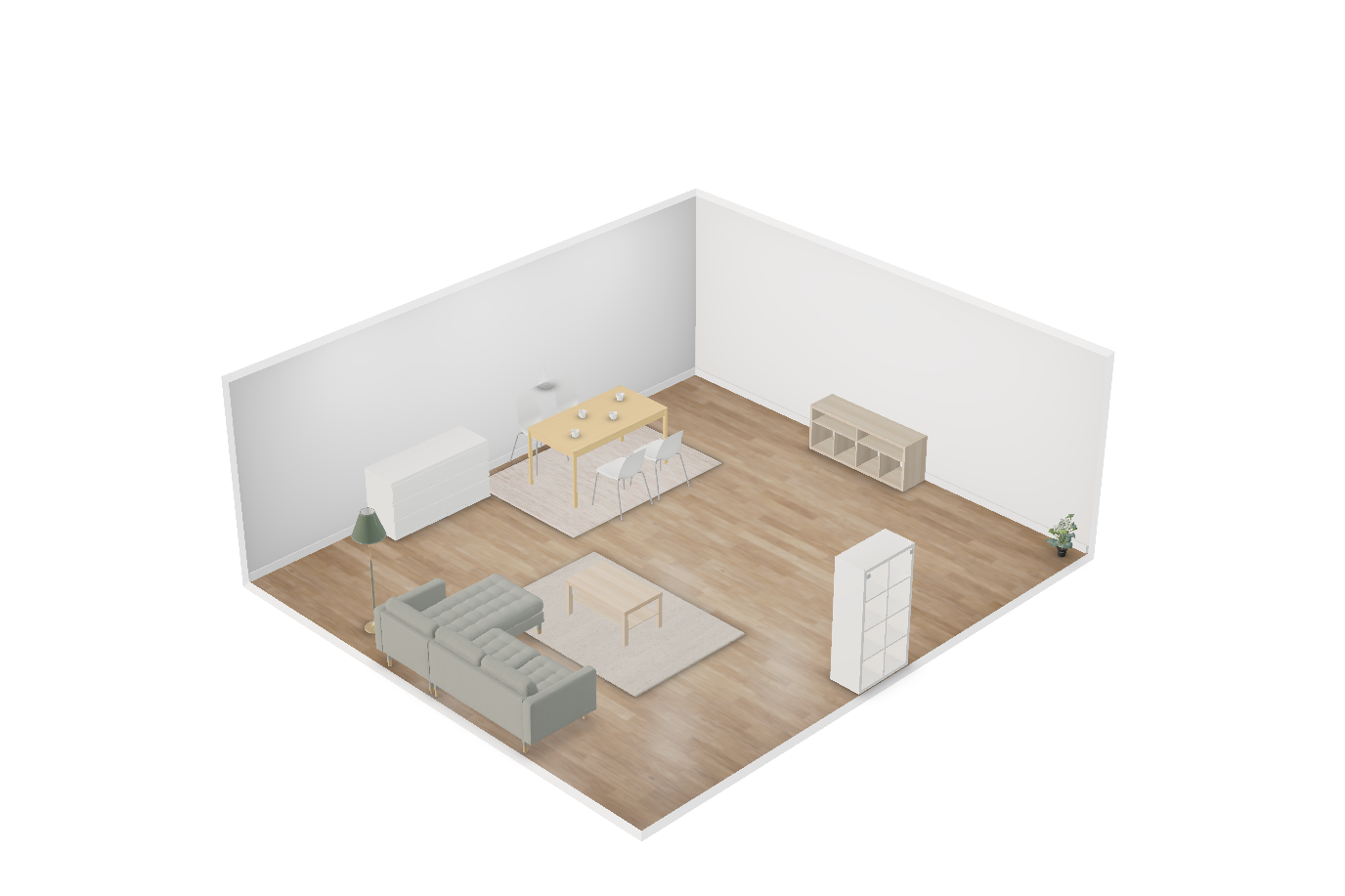} \\
  \end{tabular}
  \imgcopyright
  \caption{Qualitative comparison on three rectangular rooms. Columns share the
    same room and item set; rows are \textsc{Holodeck}, \textsc{LayoutVLM}, and
    \textsc{PolyLayout}. \textsc{PolyLayout} keeps functional groups intact and
    avoids the overlaps and gaps visible in the baselines.}
  \label{fig:qualitative}
\end{figure*}

\paragraph{\textsc{PolyLayout} resolves the trade-off.}
By resolving collisions on the discretised plan \emph{after} VLM
macro-routing, our method decouples the two axes instead of trading
them off: it attains the highest plausibility ($2.74$) while keeping
success and in-bounds perfect ($100\%$, $1.000$) and placement
near-complete ($0.973$). \Cref{fig:qualitative} shows representative
layouts. Under paired testing the plausibility gain is significant
against \textsc{LayoutVLM} ($+0.25$, $p\!<\!0.001$) and marginal
against \textsc{Holodeck} ($+0.11$, $p\!\approx\!0.05$); the blind
human study (\cref{sec:human-eval}) independently confirms the
ranking. Absolute performance still leaves headroom. The best
plausibility ($2.74$) sits just below the rubric's ``coherent and
functional'' tier ($3$), consistent with our framing of generated
layouts as starting points for customer refinement; inspecting
low-scoring layouts, the dominant flaw traces to the product-type
basis of fine placement: the taxonomy maps a fixed set of product
types onto bundle roles (\cref{sec:fine_placement}), so items of
uncovered types are placed standalone and can drift away from their
natural functional group (\eg, a chair whose specific product type
maps to no seating role is routed independently of the
bundle), whereas broken groups are absent. On latency, \textsc{PolyLayout} is comparable to
\textsc{Holodeck} ($70$s vs.\ $62$s) and roughly $13\times$ faster
than \textsc{LayoutVLM} ($907$s), though further reduction would be
needed for real-time interaction.

\begin{table*}[t]
  \centering
  \caption{Main results on rectangular rooms without openings. $\uparrow$/$\downarrow$
    mark whether higher/lower is better; Plaus.\ is a $1$--$4$ Gemini~3.1~Pro judge
    score. In-b.\ and Plaus.\ use the $N{=}255$ conditions non-empty for all methods
    (\textsc{Holodeck} returns $9$ empty rooms); other metrics use all $264$, as shown
    per column. Best in \textbf{bold}.}
  \label{tab:main}
  \begin{tabular}{@{}lccccc@{}}
    \toprule
    Method & Succ.\,(\%)\,$\uparrow$ & In-b.\,$\uparrow$ & Plaus.\,$\uparrow$ &
    Place.\,$\uparrow$ & Lat.\,(s)\,$\downarrow$ \\
     & {\scriptsize$N{=}264$} & {\scriptsize$N{=}255$} & {\scriptsize$N{=}255$} &
       {\scriptsize$N{=}264$} & {\scriptsize$N{=}264$} \\
    \midrule
    \textsc{Holodeck}  & 96.6 & \textbf{1.000} & 2.63 & 0.827 & \textbf{62.4} \\
    \textsc{LayoutVLM} & \textbf{100.0} & 0.996 & 2.49 & \textbf{0.989} & 906.8 \\
    \textbf{\textsc{PolyLayout}} & \textbf{100.0} & \textbf{1.000} & \textbf{2.74}
    & 0.973 & 70.5 \\
    \bottomrule
  \end{tabular}
\end{table*}

\subsection{Human Evaluation}
\label{sec:human-eval}

Since plausibility score is produced by a VLM judge, we test whether these scores are
grounded in what is actually rendered rather than hallucinated. Five annotators each judged the same $60$ randomly sampled conditions (fixed seed), with the three methods' renderings shuffled into blind, randomized slots; empty-room conditions are excluded as in the quality metrics, leaving $58$ conditions ($290$ votes). \textsc{PolyLayout} is preferred in $61\%$ of votes (\cref{tab:human}) and is the first choice of every one of the five raters, despite only fair inter-rater agreement (Fleiss' $\kappa=0.26$) that reflects genuinely differing tastes. Crucially, on $72\%$
of conditions the layout humans collectively prefer is the one the judge scores
highest, well above the one-third chance level, indicating the plausibility
signal reflects perceived quality rather than judge hallucination. Forced-choice preference and mean plausibility measure different things (humans penalise \textsc{Holodeck}'s sparsely-filled rooms more sharply than the pointwise judge), so we do not claim full-ranking agreement.

\begin{table}[t]
  \centering
  \caption{Blind human preference ($5$ raters, $290$ votes, empty-room conditions
    excluded). Best in \textbf{bold}.}
  \label{tab:human}
  \begin{tabular}{@{}lcc@{}}
    \toprule
    Method & Votes & Share (\%) \\
    \midrule
    \textbf{\textsc{PolyLayout}} & \textbf{177} & \textbf{61.0} \\
    \textsc{LayoutVLM} & 62 & 21.4 \\
    \textsc{Holodeck}  & 38 & 13.1 \\
    Tie & 13 & 4.5 \\
    \bottomrule
  \end{tabular}
\end{table}

\subsection{Capability Study: Openings and Non-Rectangular Plans}
\label{sec:generalization}

\begin{figure*}[t]
  \centering
  \setlength{\tabcolsep}{1pt}
  \begin{tabular}{ccc}
    \includegraphics[width=.32\linewidth]{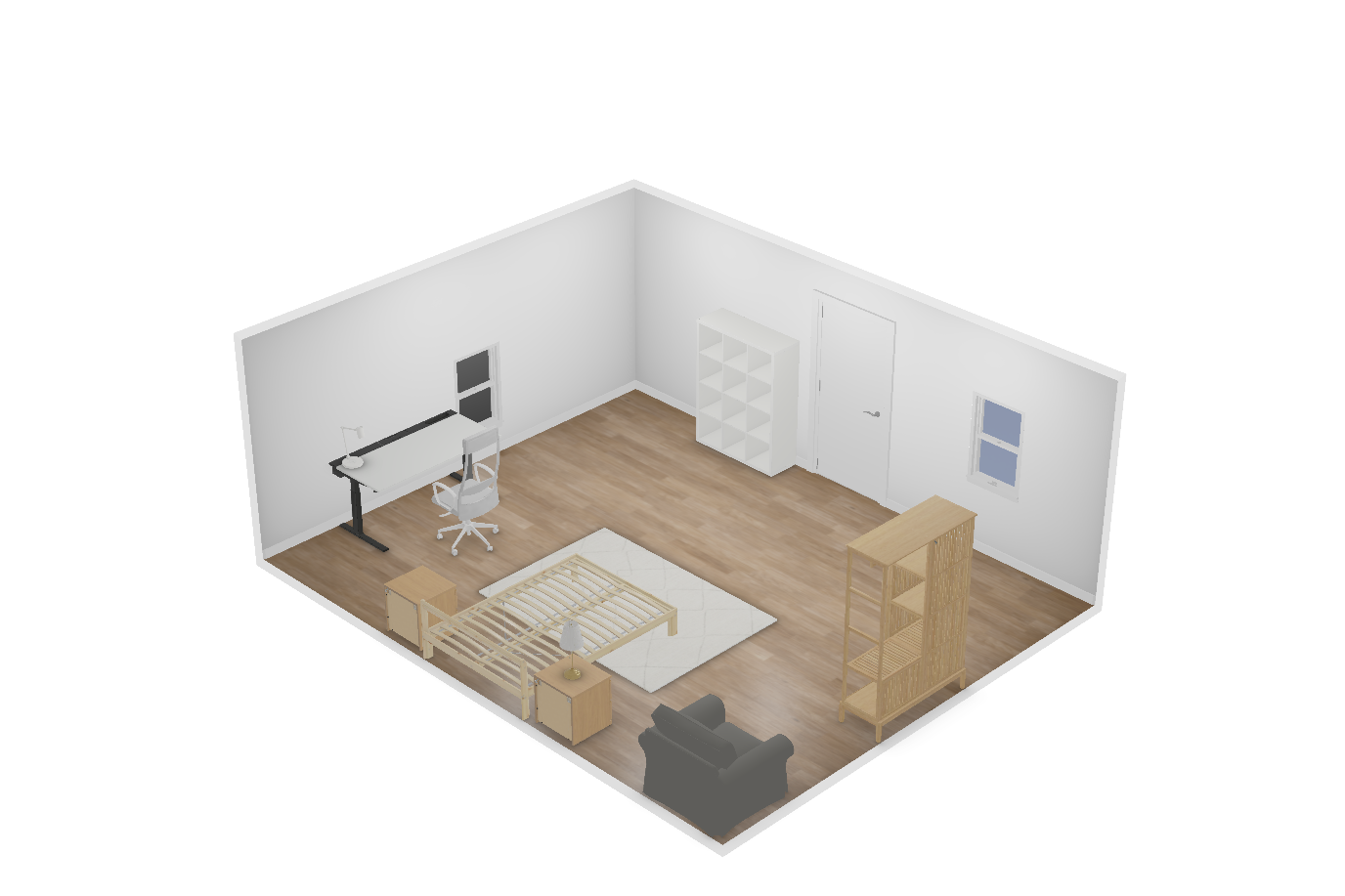} &
    \includegraphics[width=.32\linewidth]{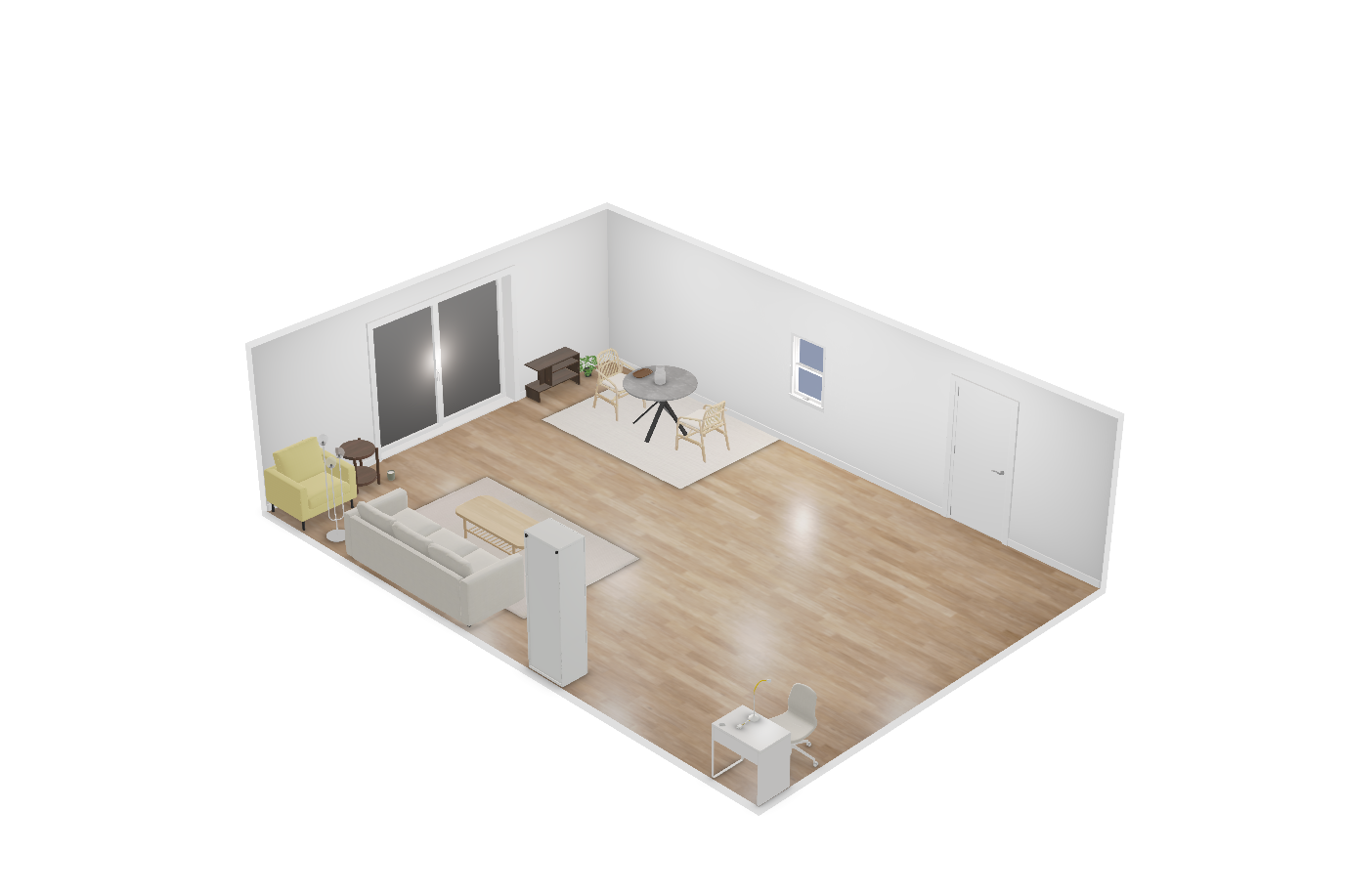} &
    \includegraphics[width=.32\linewidth]{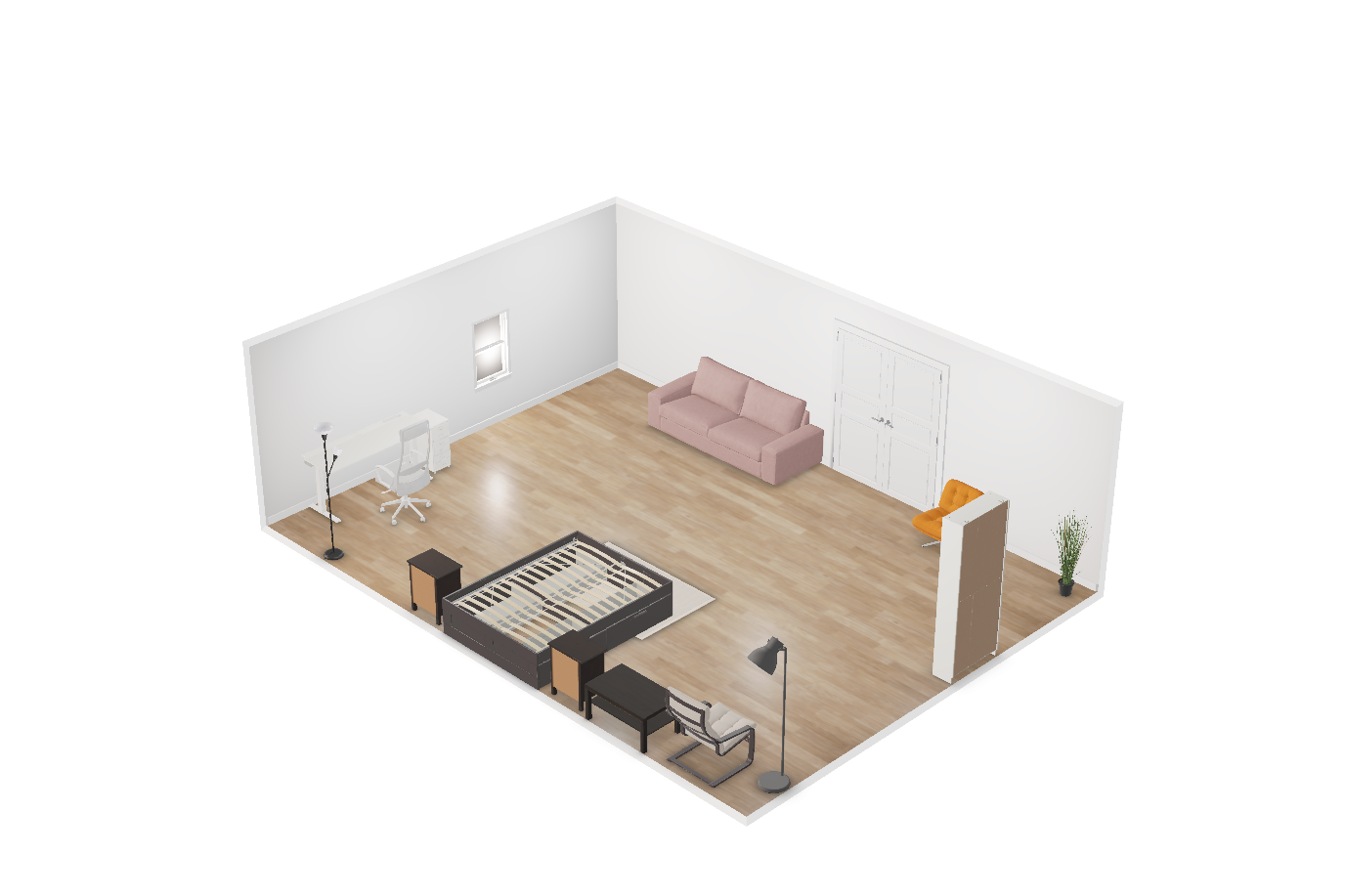} \\
  \end{tabular}
  \imgcopyright
  \caption{\textsc{PolyLayout} on rectangular rooms with doors and windows.
    Furniture clears the openings and stays in-bounds.}
  \label{fig:gen-openings}
\end{figure*}

\begin{figure*}[tbp]
  \centering
  \includegraphics[width=.32\linewidth]{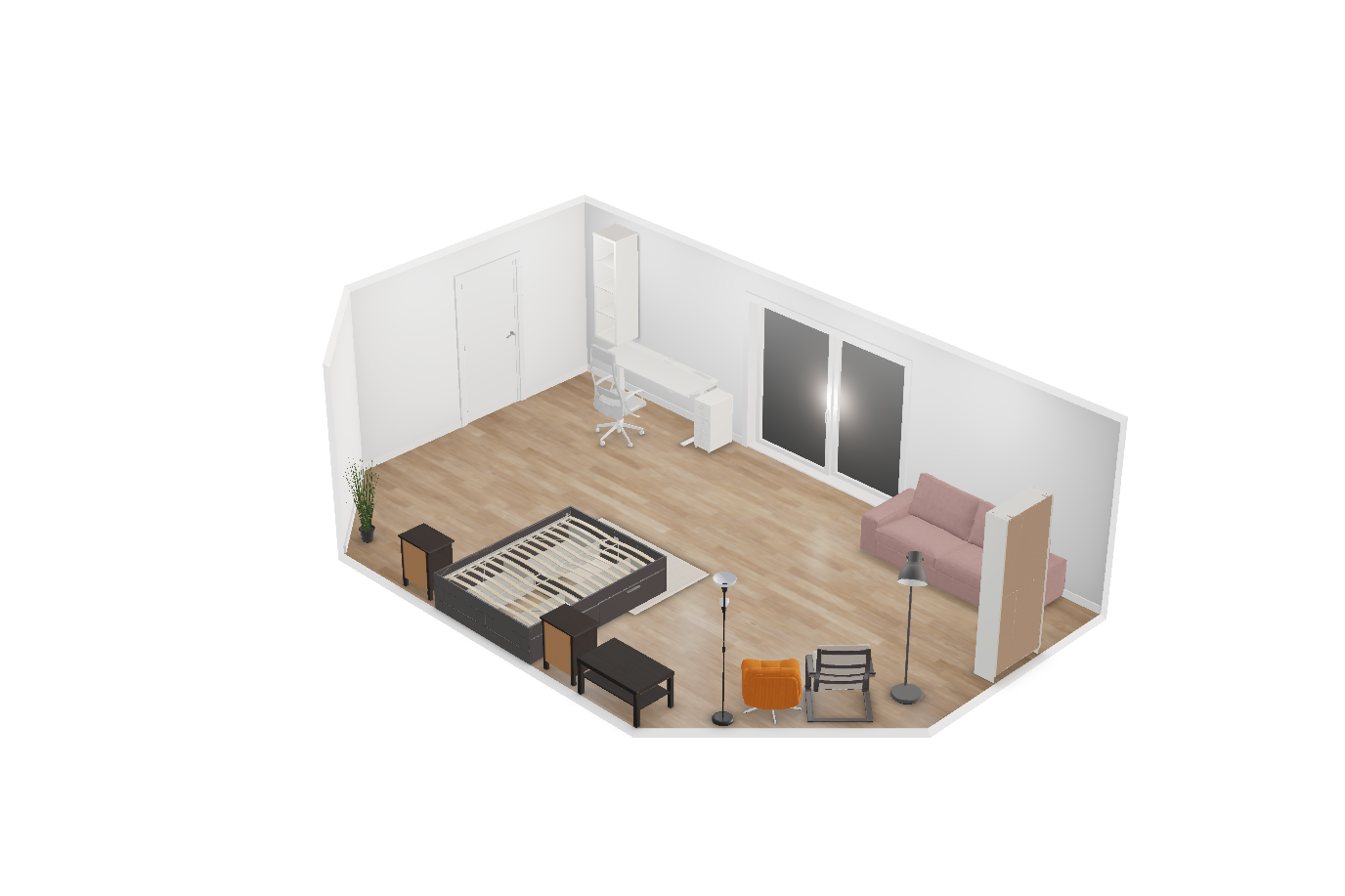}\hfill
  \includegraphics[width=.32\linewidth]{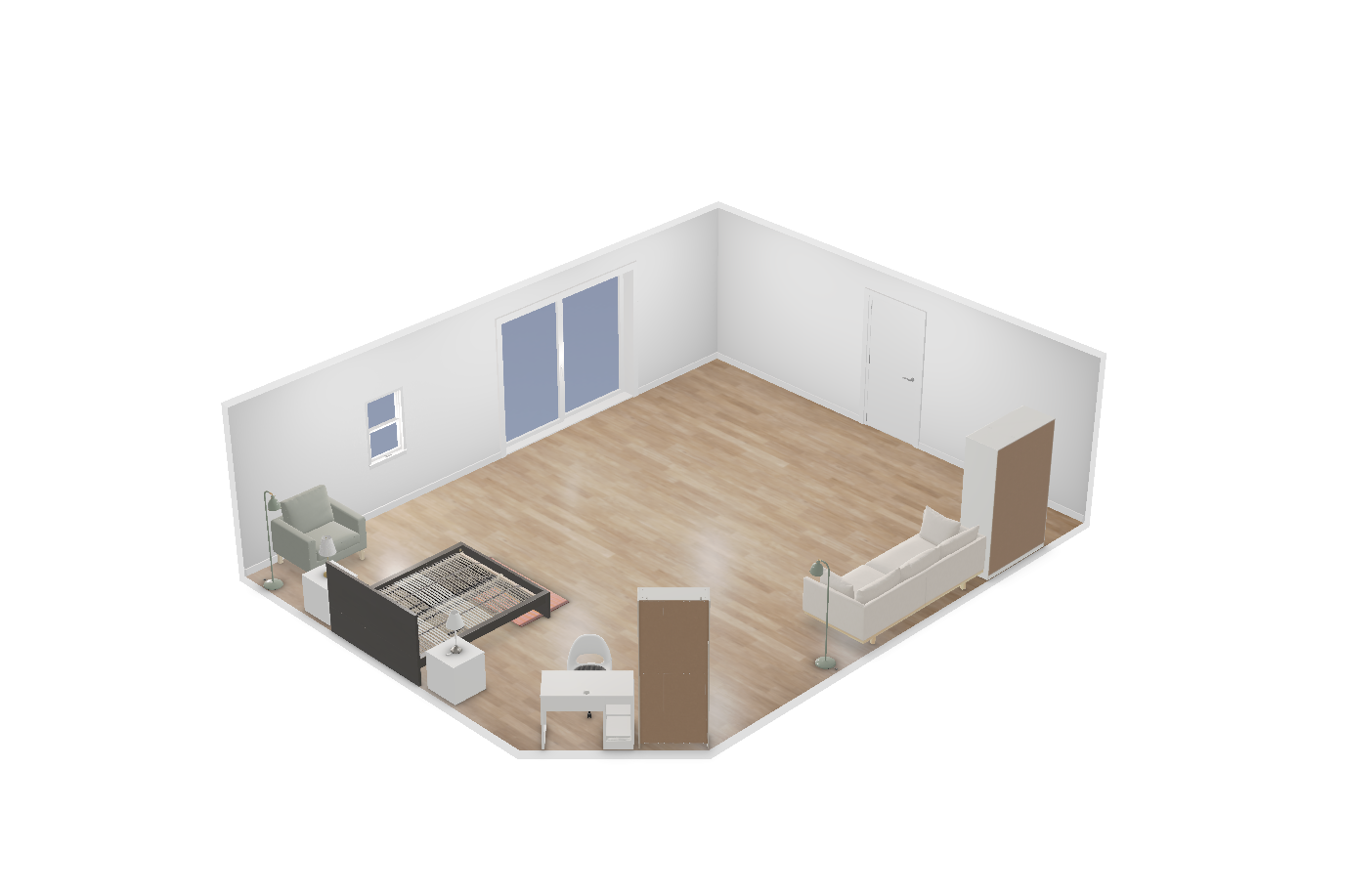}\hfill
  \includegraphics[width=.32\linewidth]{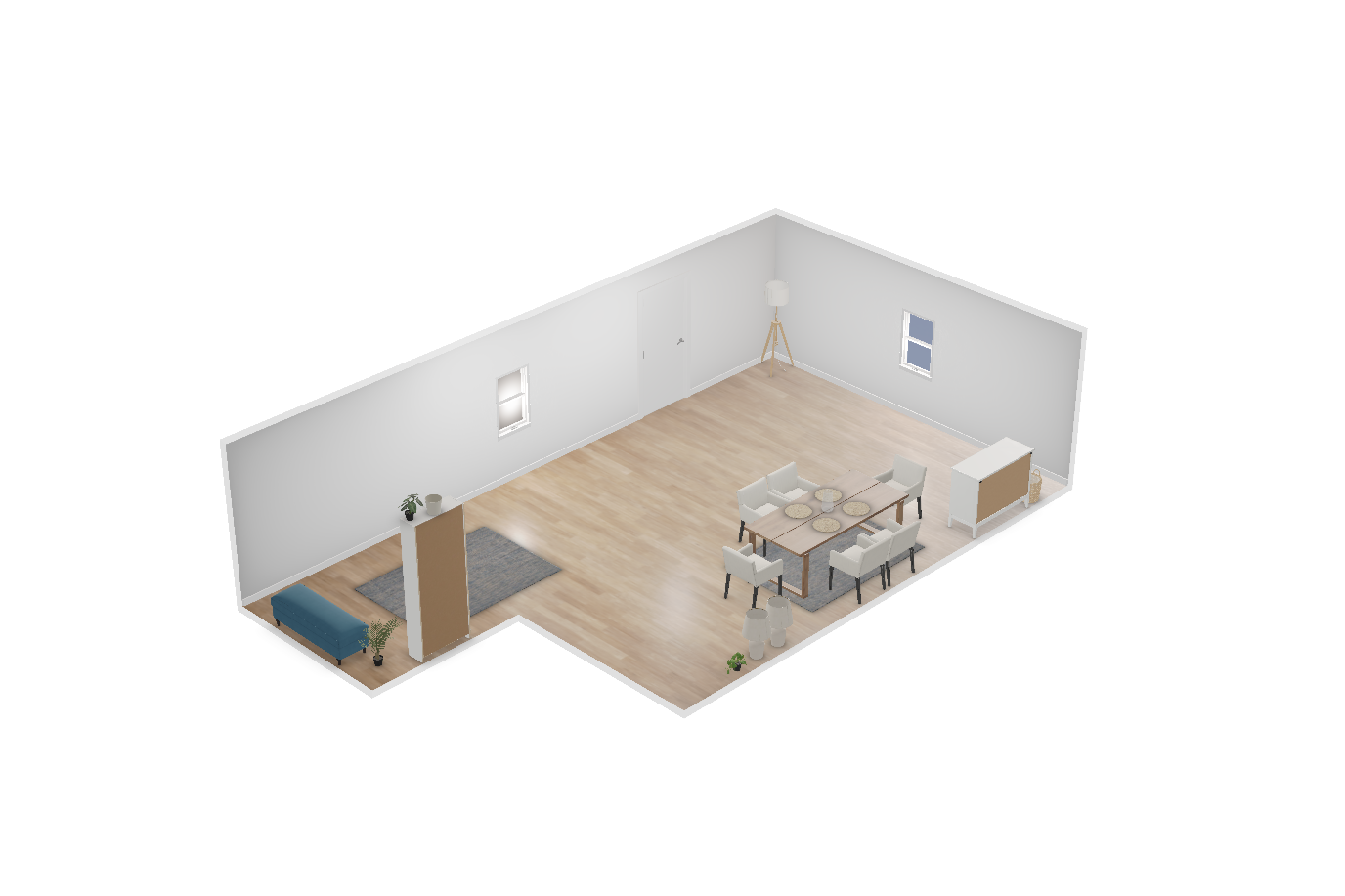}\\[2pt]
  \begin{minipage}{.66\linewidth}
    \centering
    \includegraphics[width=.485\linewidth]{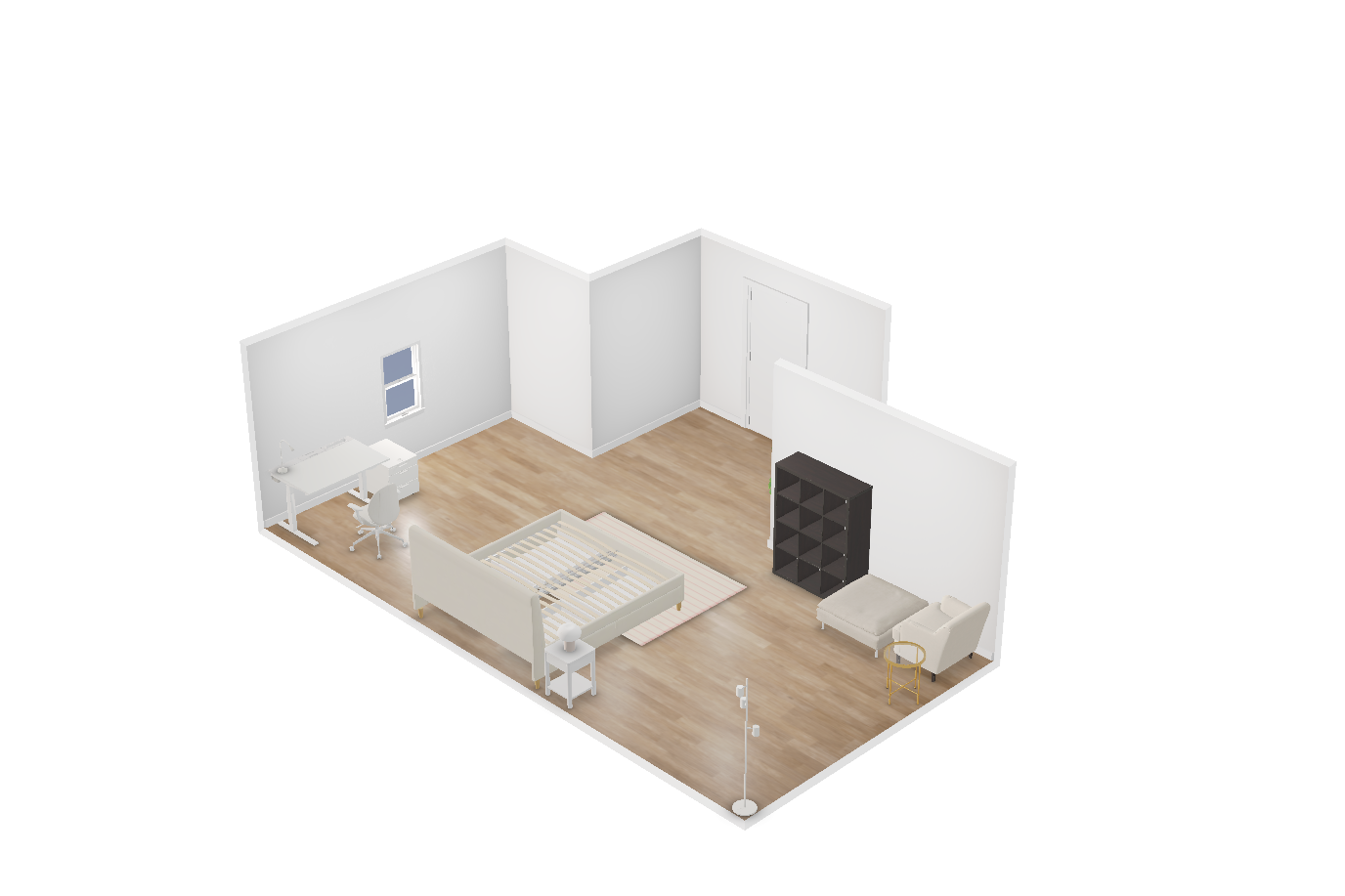}\hfill
    \includegraphics[width=.485\linewidth]{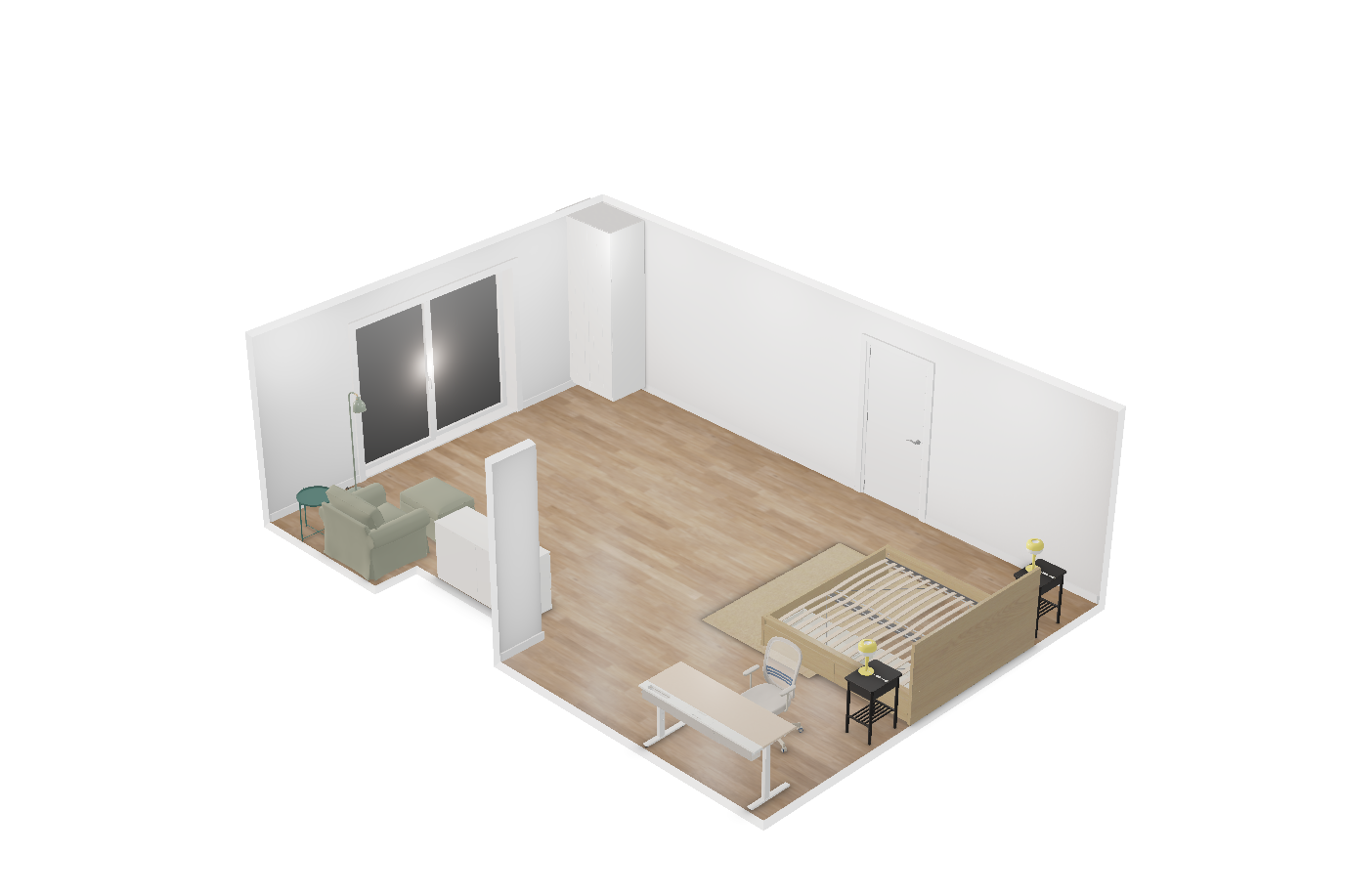}
  \end{minipage}
  \imgcopyright
  \caption{\textsc{PolyLayout} on the five non-rectangular boundaries
    (L, T, U, beveled, cut). Placements follow the irregular walls and openings
    without exiting the polygon.}
  \label{fig:gen-nonrec}
\end{figure*}

With inventories and algorithm fixed, we run \textsc{PolyLayout} on progressively harder floor plans outside the scope of the controlled comparison: the three rectangular rooms without openings, three different
rectangular rooms with door/window openings, and five non-rectangular (beveled/cut/L/T/U) boundaries. As discussed in the setup, neither baseline can condition on fixed openings or arbitrary polygons, and adapting their generative modules would yield a comparison reflecting our modifications rather than the original methods; we therefore read these results as a capability study of \textsc{PolyLayout} alone: descriptive rather than comparative. Difficulty rises at each step, yet performance degrades only gently: the hard boundary constraint is never violated (in-bounds $=1.000$ throughout) and average completeness stays $\geq0.95$, while perceptual plausibility slips just $2.73\!\to\!2.63\!\to\!2.48$ (\cref{tab:generalization}). The solver stays inside the polygon and
fills the room in every setting, so the pipeline degrades gracefully rather than breaking as the geometry hardens. \Cref{fig:gen-openings} and \cref{fig:gen-nonrec} show example layouts with openings and on non-rectangular boundaries, respectively. 

\begin{table}[t]
  \centering
  \caption{\textsc{PolyLayout} across floor-plan settings (descriptive; same
    metrics as \cref{tab:main}). Plaus.\ here averages over all $N$
    conditions rather than the matched non-empty subset of
    \cref{tab:main}, hence $2.73$ vs.\ $2.74$ in the first row.}
  \label{tab:generalization}
  \begin{tabular}{@{}lccccc@{}}
    \toprule
    Setting & Rooms & $N$ & In-b.\,$\uparrow$ & Plaus.\,$\uparrow$ & Place.\,$\uparrow$ \\
    \midrule
    Rect., no openings        & 3 & 264 & 1.000 & 2.73 & 0.973 \\
    Rect., +doors/windows     & 3 & 264 & 1.000 & 2.63 & 0.952 \\
    Non-rect., +doors/windows & 5 & 440 & 1.000 & 2.48 & 0.964 \\
    \bottomrule
  \end{tabular}
\end{table}

\subsection{Ablation Study}
\label{sec:ablation}

To isolate the contribution of each stage of our pipeline, we ablate three core
components and re-evaluate on the full benchmark under identical conditions:
\textbf{A1 (w/o clustering)} removes the grouping stage that forms
functional zones (\eg a coordinated sofa--rug--coffee-table group);
\textbf{A2 (w/o RAG plan)} removes the retrieval-augmented design plan while
keeping the module that grounds suggestions into placement functions;
and \textbf{A3 (w/o solver)} removes the dependency ordering and the geometric
repair fallback. Results are reported in \cref{tab:ablation}.

\begin{table}[t]
\centering
\caption{Ablation of the three core components. Metrics as in \cref{tab:main}; $\sigma$ is the plausibility standard deviation. Values are means with $95\%$ confidence intervals. $^{\ast}$ marks a statistically significant change
from the Full model (paired $t$-test, Bonferroni-corrected over the
$3\!\times\!3$ reported comparisons, $p<5.6\times10^{-3}$); n.s.\ otherwise.
In-bounds is $1.000$ for all variants except A3, where removing the solver lets a small tail of placements exit the room. All values are computed over the full $264$
conditions.}
\label{tab:ablation}
\setlength{\tabcolsep}{6pt}
\begin{tabular}{lcccc}
\toprule
Variant & In-b.\ $\uparrow$ & Plaus.\ $\uparrow$ & $\sigma$ $\downarrow$ & Place.\ $\uparrow$ \\
\midrule
Full model            & $1.000$          & $\mathbf{2.73}_{\pm.08}$          & $0.69$ & $\mathbf{0.973}_{\pm.008}$ \\
\midrule
A1 \emph{w/o clustering} & $1.000$        & $2.24_{\pm.08}^{\ast}$           & $0.66$          & $0.923_{\pm.012}^{\ast}$ \\
A2 \emph{w/o RAG plan}   & $1.000$        & $2.59_{\pm.08}^{\text{ n.s.}}$  & $0.70$          & $0.963_{\pm.007}^{\text{ n.s.}}$ \\
A3 \emph{w/o solver}     & $0.968_{\pm.007}^{\ast}$ & $2.41_{\pm.12}^{\ast}$ & $0.97$          & $0.900_{\pm.014}^{\ast}$ \\
\bottomrule
\end{tabular}
\end{table}

The three components degrade complementary axes rather than a single score.
\textbf{Clustering (A1)} drives overall layout quality: its removal produces the
largest and most significant plausibility drop and shifts the entire score
distribution downward (median $3\!\to\!2$), consistent with the loss of coherent
functional zones. \textbf{The solver (A3)} instead governs robustness: it leaves
the median layout unchanged (median $3\!\to\!3$) yet nearly doubles the score
variance ($\sigma\!:0.69\!\to\!0.97$) and is the only variant to break the
in-bounds ceiling ($1.000\!\to\!0.968$): removing dependency ordering and the
repair fallback does not lower the typical layout but introduces a heavy tail of
geometrically infeasible ones. \textbf{The RAG design plan (A2)} shows the smallest effect: its
score differences are not statistically significant after correction,
indicating that the retrieved guidelines polish an already-viable
layout rather than determine its feasibility. Its practical value
lies elsewhere: retrieval is what lets a retailer's own design
practice, such as walkway clearances and seating distances, shape
the layout instead of the VLM's generic priors, and the same
interface can carry user-stated preferences toward interactive
refinement.

\FloatBarrier

%% file: sections/conclusion.tex
\section{Conclusion}
\label{sec:conclusion}

We presented \textsc{PolyLayout}, a hierarchical layout generator that decouples semantic arrangement (VLM-guided macro-routing grounded in design knowledge) from a deterministic geometric solver that enforces physical
feasibility directly on the floor-plan polygon, including arbitrary boundaries and openings.

On a retail benchmark of real IKEA home furnishing products in real rooms with the Gemini~2.5~Flash backbone, \textsc{PolyLayout} attains the best plausibility and perfect in-bounds placement at CPU-level latency, a ranking confirmed by both a VLM judge and a blind human study; the same pipeline extends without modification to rooms with openings and to non-rectangular L/T/U/beveled/cut plans, a regime neither baseline natively supports.

Our evaluation is limited to single rooms from the IKEA home furnishing product catalog, the cross-setting capability results are descriptive rather than controlled since each setting uses different rooms, all geometric conclusions rest on
eleven rooms (six rectangular rooms, five non-rectangular rooms; condition counts scale with inventories and repeated runs rather than room variety), and no existing baseline conditions on fixed openings or polygonal boundaries, leaving the capability study without a comparator. Future work includes multi-room and whole-home layouts, evaluation across multiple VLM backbones, learning the solver's cost terms from human preferences, and controlled
comparisons as constraint-conditioned baselines emerge.